\documentclass[authoryear,3p,10pt]{elsarticle}

\usepackage{subfigure}
\usepackage{graphicx}
\usepackage{gensymb}
\usepackage{placeins}
\usepackage{amsmath}
\usepackage{amssymb}
\usepackage{amsthm}
\usepackage{afterpage}

\usepackage{hyperref}

\journal{ISPRS Journal of Photogrammetry and Remote Sensing}

\def\BibPath{.}  

\begin{document}

\begin{frontmatter}

\title{
Prioritized Multi-View Stereo Depth Map Generation Using Confidence Prediction
}

\author[mymainaddress]{Christian Mostegel\corref{mycorrespondingauthor}}
\cortext[mycorrespondingauthor]{Corresponding author}
\ead{mostegel@icg.tugraz.at}

\author[mymainaddress]{Friedrich Fraundorfer}
\ead{fraundorfer@icg.tugraz.at}

\author[mymainaddress]{Horst Bischof}
\ead{bischof@icg.tugraz.at}

\address[mymainaddress]{Institute for Computer Graphics and Vision, Graz University of Technology, Austria}

\begin{abstract}
 In this work, we propose a novel approach to prioritize the depth map computation of  multi-view stereo (MVS)
to obtain compact 3D point clouds of high quality and completeness at low computational cost.
Our prioritization approach operates before the MVS algorithm is executed 
and consists of two steps.
In the first step, we aim to find a good set of matching partners for each view.
In the second step, we rank the resulting view clusters (i.e. key views with matching partners)
according to their impact on the fulfillment of desired quality parameters such as completeness, ground resolution and 
accuracy.
Additional to geometric analysis,
we use a novel machine learning technique for training a confidence predictor.
The purpose of this confidence predictor is to
estimate the chances
of a successful depth reconstruction for each pixel in each image
for one specific MVS algorithm
based on the RGB images and the image constellation.
The underlying machine learning technique does not require any ground truth or manually labeled data for training,
but instead adapts ideas from depth map fusion for providing a supervision signal. 
The trained confidence predictor allows us to evaluate the quality of image constellations
and their potential impact to the resulting 3D reconstruction 
and thus builds a solid foundation for our
prioritization approach.
In our experiments,
 we are thus able to reach more than 70\% of the maximal reachable quality fulfillment using only 5\% of the available
images as key views.
For evaluating our approach within and across different domains, 
we use two completely different scenarios,
i.e. cultural heritage preservation and reconstruction of single family houses.

\end{abstract}


\end{frontmatter}
%

\section{Introduction}

\begin{figure}[t]
  \centering
\includegraphics[width=1\columnwidth]{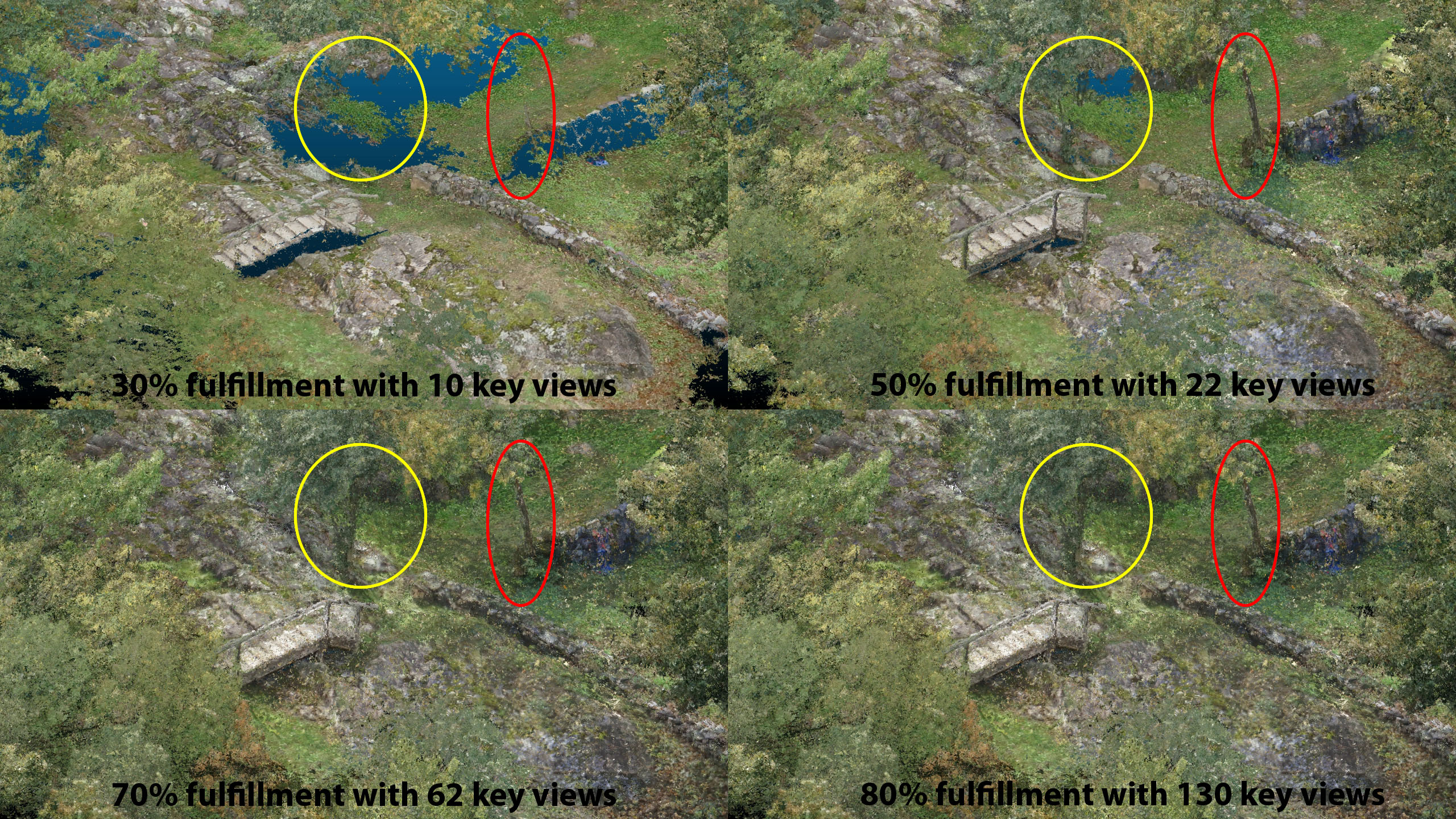} 
    
  \caption{View Cluster Prioritization. Our approach allows us to prioritize/rank view clusters (i.e. key views with matching partners)
  such that a highly complete and accurate point cloud can be obtain with a very small fraction of the available images.
  Here, we show the point clouds from the raw depth maps of the view clusters (with 11 matching partners) ranked with our approach 
  after reaching 30\%,50\%,70\% and 80\% of the maximal achievable quality fulfillment (i.e. completeness with respect to a desired ground sampling distance and accuracy of 1cm).
  Already with 50\% fulfillment and only 22 key views (i.e. 1.8\% of the available images), most parts of this complex scene are already contained in the reconstruction (red ellipse)
  and only a small part is missing (yellow ellipse).
  With 70\% fulfillment, even strongly occluded parts such as tree trunks (see ellipses)
  are contained in the point cloud, although this point cloud is computed from only 62 key views (i.e. 5\% of the available images).
  For going from 70\% to 80\% fulfillment, the number of necessary key views already has to be more than doubled,
  however, the visual difference between those two point clouds is nearly imperceptible (see also the supplemented video).
 }
  \label{fig:teaser}
\end{figure}

In this work, we aim to improve the efficiency of  multi-view stereo (MVS) approaches based on depth maps.
This type of approach is very popular (e.g. \citep{goesele07,rothermel12,zheng14,galliani15,schoennberger16}) as
it is inherently parallelizable and delivers state-of-the-art results.
One drawback of such approaches is that they typically generate one depth map per image in the dataset.
For modern cameras, this means that 3D points in the order of $10^7$ are created
per image.
With a few hundred images this leads to billions of points, that
have to be stored, visualized and/or handled by subsequent processing steps
such as depth map fusion or surface reconstruction.
In this work, we propose a way to
significantly reduce the number of generated points,
while at the same time preserving the reconstruction accuracy and completeness to a large extent (see Figure~\ref{fig:teaser}).
This does not only speed up the depth map computation process,
but also significantly reduces the load for all subsequent steps.

The key to our improved efficiency is a combination of geometric reasoning based on a preliminary scene reconstruction and 
machine learning to represent important properties of MVS that cannot be geometrically modeled.
These unmodeled properties stem from the fact that MVS is an ill-posed task 
and to solve this task all MVS algorithms have to rely on a combination of some similarity measure 
(e.g. Census Transform or Normalized Cross Correlation) with 
a set of reasonable assumptions about the scene.
The most popular assumptions include visual saliency, local planarity and 
a static scene.
While these assumptions work well in many environments, there are also many environments where parts of the assumptions do not hold.
Thus, MVS reconstructions very often contain outliers and/or fail to reconstruct 
certain objects completely.

In this work, we use our unsupervised machine learning framework~\citep{mostegel16b,mostegel16} to predict these failures
and help us to reduce the number of key views (i.e. the image for computing a depth map) and necessary matching partners.
Our method consists of two main steps.
First, we select suitable matching partners for each view.
Second, we prioritize/rank the resulting view clusters (i.e. views with matching partners) depending on their
impact on a quality fulfillment function.
This quality fulfillment function respects important photogrammetric parameters, such a ground resolution and 
3D uncertainty, together with the scene coverage.
The confidence prediction supports this whole process and 
allows us to obtain this ranking without having to execute the actual MVS algorithm within the ranking procedure.
We formulate this quality fulfillment function as monotone submodular function and 
optimize this function with our ranking procedure in a greedy fashion.
Although the overall problem is still NP-hard (as it includes the NP-hard maximum coverage problem), this formulation gives us strong optimality guarantees 
in the function space~\citep{nemhauser1978analysis}.

This formulation has many advantages.
First of all, the computed quality fulfillment function  
provides the opportunity to decide how many view clusters are necessary to obtain a certain quality fulfillment 
level.
Thus, the operator can either choose to reconstruct the $n$ best view clusters and has a estimation
of the expect level of fulfillment or can simply query how large $n$ should be to reach a certain level.
The second advantage is that the inherent parallelism of MVS based on depth maps is maintained
as our ranking procedure happens before executing the MVS reconstruction step.
Third, the overall efficiency of the MVS reconstruction step can be significantly improved without 
changing the MVS algorithm itself.
Thus, we were able to obtain a quality fulfillment (i.e. completeness with respect to a desired resolution and accuracy) of 70\% with only 5\% of the available view clusters.
This leads to a speed up factor of approximately 10 and a complexity/memory reduction factor of approximately 20 for the resulting point cloud
without losing much information.

\section{Related Work}
\label{ss:ic_related_work}

Our prioritization approach is related to two different research areas, which are namely:  Matching partner selection and Next-Best View (NBV) planning.
In the following, we discuss the relation to these two interwoven areas.

\paragraph{Matching Partner Selection}
Most MVS approaches based on depth maps formulate some kind of heuristic to select the $k$ best matching partners
for each key view to increase the efficiency of MVS.
The heuristics for matching partner selection strongly depend on how the images are acquired
(structured versus unstructured)
and the requirements of the MVS algorithm.
If the images are acquired in a regular grid, the $k$ closest images are a natural choice to maximize the completeness.
For more unstructured settings, 
the connectivity in the sparse reconstruction (i.e. how many sparse 3D points are shared between two cameras)
is typically a more reliable cue to determine if the dense MVS matching step will work or not.
To avoid that images with insufficient parallax are chosen as matching partners, \cite{goesele07} combine the connectivity with geometric constraints in a greedy fashion.
Their formulation down-weights connections (shared features) with a triangulation angle below $10^\circ$ and
dissimilar scale.
Additionally to these two terms, \cite{bailer12} also add a coverage term, which favors 
connections that have not been covered by other selected images.
\cite{shen13} use a formulation without connectivity only based on the geometric constraints on
the triangulation angle and the distance between images.
For very small datasets where all images nearly see the same part of the scene (as in the DTU dataset~\citep{aanaes2016large}),
also random selection of matching partners can lead to good results~\citep{galliani15}.
Of all formulations mentioned above, \citep{bailer12} seems to be the closest related formulation to our approach.
Similar to their approach, we also use the connectivity to reduce the set of potential matching partners.
But instead of using hand-crafted models for the triangulation angle,
our approach uses the learned weighting function (encoded as the MVS confidence).
Another difference is that we use a surface mesh to evaluate geometric constraints
such as triangulation angle, scale and coverage.
In contrast to only using sparse points, a surface mesh provides the opportunity for geometric balancing.
This means that instead of having a very unbalanced set of sparse points (with many points in richly textured parts and
no points in poorly textured parts), the triangle distribution of the mesh can be easily balanced over the whole scene.
This can be important for low texture structures, e.g. building facades, where very often no sparse points 
are extracted, whereas the micro-texture can still contain sufficient information for obtaining an accurate MVS reconstruction.

\paragraph{Next-Best View Planning}
Next-best view (NBV) planning, view planning, sensor placement,
path planning for reconstruction and/or coverage, visual inspection and exploration are all closely related topics.
They all have to answer the questions "what parts of the scene are already sufficiently covered?"
and "how can I best improve this coverage?".
In this sense, they are all related to the art gallery problem~\citep{o1987art} or the maximum coverage problem~\citep{feige98},
which both have been shown to be NP-hard.

Thus, researchers from the communities of photogrammetry, robotics and computer vision
have developed and are still developing methods which are honed to very specific tasks
exploiting all available domain specific knowledge to conquer this challenging task
with one specific actuator/sensor setup.
Some works (e.g. ~\citep{fan16,hossein14a,hossein14b,trummer10nbv}) thus focus on the reconstruction
of small scale scenes in laboratories.
For this kind of approaches we refer the reader to a recent review article of \cite{karaszewski16}.
With the increasing availability of robotic platforms, also the research in the field of structure inspection drastically increases
(e.g.~\citep{hollinger12vp_underwater,alexis15,Bircher2016b,galceran15}). 
A review of current approaches on this topic can be found in~\citep{almadhoun16}.
This field is also closely related to coverage search (e.g.~\citep{dornhege13,galceran13}).
For photogrammetry, there also exist several works on view planning for UAVs
(e.g~\citep{mostegel16b,martin16,schmid12view_planning_mav,hoppe12nbv,jing16ICARCV,jing16IROS}).
While many aspects are shared with view planning, the problem tackled in this work also shows one important
difference: In our case the set of images is fixed. On the one hand, this is an advantage as
subtasks such as path planning do not need to be handled anymore. On the other hand, if the image constellations
are suboptimal then there is no possibility to improve the situation.

From all the works mentioned above,
we see the works of \cite{hornung08} and \cite{mendez16} as most related to our task.
Both approaches try to obtain a maximally complete reconstruction from a given image set
by iteratively updating the geometry estimate.
For updating the geometry, \cite{hornung08} use the complete PMVS algorithm~\citep{furukawa10cmvs} after each added view,
while \cite{mendez16} execute \citep{weinzaepfel13} after selecting the next-best stereo pair.
We contrast from these works in several points.
First of all, our aim is not an iterative update scheme with the MVS algorithm in the loop,
but instead we aim to rank suitable view clusters (i.e. key views with matching partners) according to
their importance for the final reconstruction.
On the one hand, we are thus able to obtain a complete reconstruction
with a fraction of available view clusters.
On the other hand,
we can pre-compute the whole key view 
ranking, which allows us to maintain the natural parallelization
capabilities of depthmap-based MVS in the execution phase.
Instead of updating the geometry estimate in each iteration,
we use all measurements of the sparse reconstruction
to obtain a high quality surface mesh at a low computational cost once at the start of our algorithm.
The big advantage of this strategy is that we do not require the MVS algorithm in the loop,
but use our MVS confidence prediction to estimate the chances of a good reconstruction.
Through this formulation, we combine the advantages of NBV planning and 
depthmap-based MVS, i.e. data reduction and parallelism.
This leads to a light-weight approach, which can be easily integrated in most photogrammetric pipelines
to increase the overall efficiency.

%

\section{View Cluster Prioritization}
\label{sec:ic}
The basic element of our prioritization is a view cluster, which stands for a key view (i.e. an image used for generating a depth map)
together with its matching partners.
The aim of our view cluster prioritization is to establish a ordering of view clusters 
such that any subset of the ranked entries from entry 1 to $n$ retains a maximum of information about the scene for any $n$.
In this work, we separate the problem of view cluster prioritization into two main subproblems.

The first subproblem is matching partner selection.
For this subproblem, let us assume that we have decided that we want to generate a 
depth map for a distinct view/image, which we will further call "key view".
To generate a depth map, we require other images that observe the same part of the scene from different positions.
Choosing a subset of images for this task has a lot of impact on the quality of the resulting depth map.
On the one hand, choosing images with a large baseline to the key view will lead to depth values with high accuracy.
On the other hand, it will also make the matching task (i.e. finding correspondences between the images)
significantly harder, which will negatively influence the completeness of the depth map.
Selecting more images will be beneficial for completeness and accuracy, however,
the matching time also rises linear with the number of matching partners.
So for increasing the efficiency, the number of matching partners is typically kept as low as possible.
Thus, in this subproblem it is our aim to find the best subset of $k$ matching partners
such that $k$ can be very small and we still obtain high quality results in terms of accuracy and completeness.

The second subproblem is next-best view ranking.
For this subproblem, let us assume that we have already decided the set of matching partners for each key view.
While it is possible to generate one depth map for each view, this approach leads
to a massive amount of highly redundant data.
Thus it is our aim to rank the view clusters (key views with matching partners) such that the most 
useful view clusters can be processed first.
As by product, our approach delivers a fulfillment prediction (with respect to a desired ground resolution and 3D accuracy)
for each added view cluster.
This information makes it possible to determine how many and which view clusters are necessary to obtain a certain level
of fulfillment prior to executing the MVS algorithm itself.

\begin{figure}[t]
  \centering
\includegraphics[width=1\columnwidth]{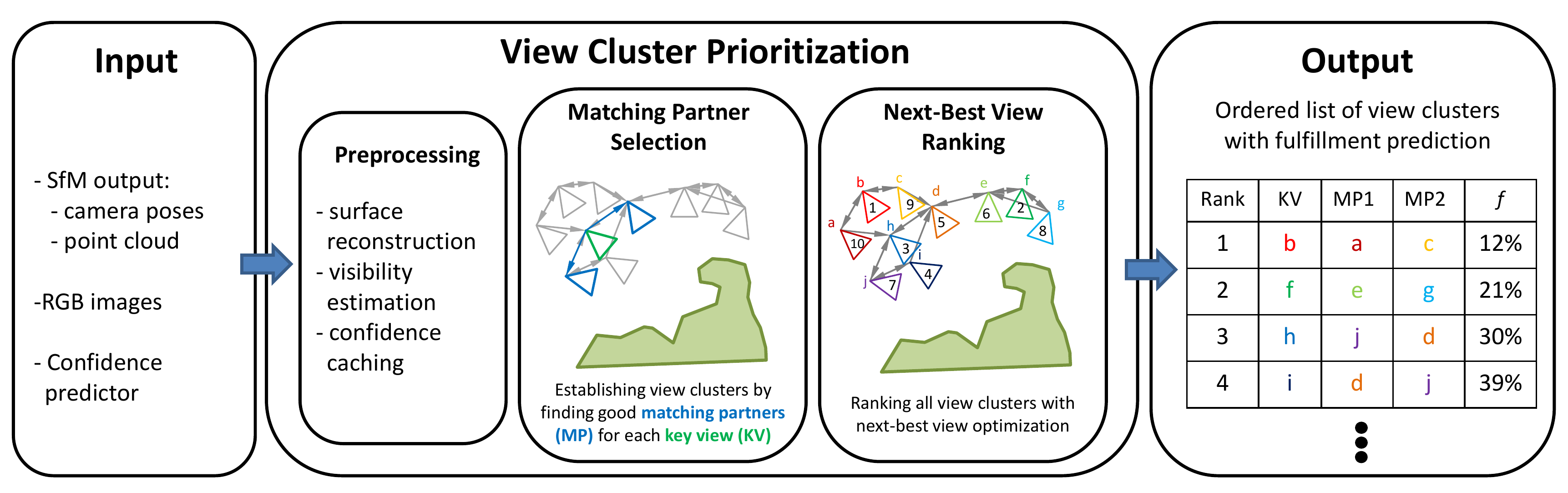} 
    
  \caption{View Cluster Prioritization. As input our approach requires the SfM output (i.e. the camera poses and the sparse point cloud),
  the original RGB images and the pre-trained confidence predictor. 
  From the SfM output, we extract a surface mesh, which is then used for visibility estimation (i.e. estimation of which camera sees which mesh triangles).
  With this information, we then precompute the MVS confidence for each image as described in~\citep{mostegel16b} and cache the predictions for each visible mesh triangle.
  Using this data, we then find good matching partners for each view, which results in a set of view clusters.
  These view clusters are then ranked by their importance for the overall reconstruction in a next-best view scheme. The output is ordered list of view clusters (i.e. key views with matching partners) together with
  a fulfillment prediction $f$ for each entry.
  }
  \label{fig:vcr_scheme}
\end{figure}

In Figure~\ref{fig:vcr_scheme}, we show an overview of the complete ranking procedure, which is 
explained in full detail in the remainder of this section.

\subsection{Preprocessing}
As input our approach requires the RGB images, a pre-trained confidence predictor \citep{mostegel16b} and the structure-from-motion output
including the corresponding camera poses and a sparse 3D point cloud.
For our metric quality constraints (ground sampling distance and 3D accuracy), we also require 
that the structure-from-motion output was transformed metrically.
In our experiments, we used automatically detectable GCPs~\citep{rumpler17} for this purpose.

From this input, we then robustly extract a surface mesh with a Delaunay triangulation of 
the sparse point cloud~\citep{labatut07delaunay} with the same parameters used in \citep{mostegel17}.
The amount of triangles of the resulting surface mesh is in the order of the number of sparse 3D points.
For view cluster prioritization, the resulting geometry complexity is typically already unnecessarily high.
Therefore, we first simplify the mesh reconstruction using an adaption of quadratic edge collapse decimation~\citep{garland97},
which terminates when 95\% of all triangle edges are above $r$ times the desired ground sampling distance.
In our experiments, we found $r = 20$ to be good value, as with this value all important 3D structures are still contained in the 
mesh, but the number of triangles is drastically reduced.
To balance the triangle size independent of the 3D topology, 
we then refine large mesh triangles by iterative sub division until all edge lengths 
are below $e$ times the desired ground sampling distance.
We found that $e = 5 \times r$ lead to a good trade-off between simplicity and balance.
Note that the surface  mesh is computed exactly once at the beginning of our approach and
is not refined with the MVS output as our prioritization approach works before the MVS algorithm is executed.
%

\subsection{Matching Partner Selection}
For computing a depth map with MVS, each key view requires a set of matching partners which observe the same scene
from different view points.
In this step of our approach, we try to find a good set of $k$ matching partners in the sense that completeness and 
accuracy are optimized simultaneously.
We say that we are aiming for a \emph{good} set (opposed to the optimal set) as even the solution space for a single key view can be extremely large.
The size of the solution space for this task is the same as for unordered sampling without replacement, i.e. the binomial coefficient $\frac{n!}{k!(n-k)!}$, 
where $n$ is the size of the potential set of matching partners.
Even for quite small numbers, like picking $k=6$ matching partners out of 
$n=23$ possibilities, the solution space is already larger than 100k.
For this purpose, we follow other works in the field (e.g.~\citep{goesele07,bailer12})
and use the connectivity between the images for a complexity reduction.
Thus, we first reduce the set of potential matching partners to the top $n$ most connected images.
We evaluated the connectivity based on the number of sparse points shared with the key view. 
From this set we then draw $y$ combinations.
As the connectivity can be a very strong cue, we also want to ensure that the most connected images are included in the drawn $y$ combinations.
Thus, we first draw all possible combinations of the $q$ most connected images, where $q$ is the 
largest set size for which the solution space is less or equal $\frac{y}{4}$ (i.e. $\frac{q!}{k!(q-k)!} \leq \frac{y}{4}$).
The rest of the $y-q$ combinations are drawn randomly from the larger set of the $n$  most connected images.
In this way at least 75\% of the combinations are drawn completely random, whereas 
we also consider the most connected combinations.
For each drawn combination of matching partners ($c^k$), we now evaluate the following four fulfillment functions in relation to the key camera $c_{key}$
and a triangle $t$ observed from $c_{key}$:

(1) The \emph{coverage} is modeled as a Boolean function $f_{cov}$, which is true if sufficient cameras ($\geq x$) observe the triangle $t$ and false otherwise.

(2) The \emph{resolution} requirement ($px/m^2$) is defined as a truncated ratio:
\begin{equation}
 f_{res} = \min\left\{ \frac{r}{r_d} , 1 \right\},
\end{equation}
where $r_d$ is the desired resolution and $r$ is the estimated resolution.
We determine $r$ by projecting the 3D triangle into $c_{key}$ and dividing the 2D triangle area 
by the 3D triangle area. The desired resolution $r_d$  can be computed from a desired ground sampling distance $g_d$ as $r_d = 1 / g_d^2$.

(3) The fulfillment of the \emph{3D uncertainty} requirement for a desired accuracy $a_d$ is defined
as:
\begin{equation}
 f_{unc} =\min\left\{  \frac{a_d}{\sqrt{u}} , 1 \right\},
 \label{eq:aia_func}
\end{equation}
Here, $u$ stands for the maximum Eigen value of the covariance matrix $\textit{Cov}_{3D}$ related
to a triangle's centroid~\cite{hartley04multiview}.

(4) The last fulfillment function is the MVS confidence for $k$ matching partners.
In~\citep{mostegel16b}, we presented a way to learn and predict this confidence without any ground truth for two matching partners.
Here we present a way to generalize the confidence prediction step to an arbitrary number of $k$ matching partners.
For this purpose, we assume that the MVS reconstruction process follows a stochastic process
and can be modeled as a combination of unary MVS confidence predictions.

Under this assumption, we can use probability theory to reason about 
 the overall probability of obtaining a successful measurement for a pixel of a key view 
 given a set of matching partners.
 For a successful measurement, most MVS approaches require successful matches to at least two matching partners.
 Thus, we formulate the overall probability of obtaining a successful measurement, as the probability
 of getting successful matches to at least two matching partners.
In~\ref{conf_extension_appendix}, 
we derive this overall probability by growing a binary probability tree.
In the following, we apply the general solution described in~\ref{conf_extension_appendix} to our specific problem.

In our problem, we are given a surface mesh with a set of triangles $T$ and a set of images $C$.
For a specific combination of a key view $c_{key}\in C$ with $k$ matching partners $c^k \in C^k$,
we want to predict the chances of successfully reconstructing the depth of the pixels in which a triangle $t \in T$ projects.

 Thus, we now end up with 
the following equation for the overall probability of a successful match considering all matching partners:
\begin{equation}
  f_{conf}(t,c_{key}, c^k) = \sum_{i=2}^k \left( (-1)^i \cdot (i-1) \cdot \sum_{c^{(i)} \in C^{(i)}_{k} }  \left( \prod_{j=1}^i f_{conf}\left(c_{key},c^{(i)}_j\right)  \right) \right) ,
  \label{eq:prob}
\end{equation}
where $C^{(i)}_{k}$ is the solution space for drawing subsets of $i$ cameras
from the available set of $k$ matching partners, $c^{(i)}$ is one of these subsets and 
$c^{(i)}_j$ is one camera of this subset and
$f_{conf}\left(c_{key},c^{(i)}_j\right)$ is the pair-wise MVS confidence for camera $c^{(i)}_j$
with the key camera $c_{key}$.
We compute this pair-wise MVS confidence by averaging the unary confidence values of the two cameras.
We extract the unary confidence value for a triange $t$ by averaging the pixel-wise confidence predictions of 
all pixels that lie inside the projection of triangle $t$ into the individual camera.
The confidence predictor is trained and executed as described in~\citep{mostegel16b}.
Note that for improved efficiency, we also precompute and store the unary confidence values for each camera-triangle pair in our preprocessing step.

Based on these four fulfillment functions,
we define the fulfillment for a triangle $t$ as:
\begin{equation}
 f(t,c_{key},c^k) = (\alpha f_{res}(t,c_{key}) + (1-\alpha) f_{unc}(t,c_{key} \cup c^k) ) \cdot f_{cov}(t) \cdot f_{conf}(t,c_{key}, c^k),
 \label{eq:triangle_fulfillment} 
\end{equation}
where $\alpha$ defines the relative weight between the resolution and uncertainty fulfillment.

With this basis function,
we now compute a combined fulfillment score for each set of drawn matching partners as
\begin{equation}
 f(c_{key},c^k) = \sum_{t\in T_{c_{key}}} f(t,c_{key},c^k) ,
 \label{eq:matching_fulfillment}
\end{equation}
where $c_{key}$ is the key camera, $c^k$ is the selected set of $k$ matching partners, the triangle set $T_{c_{key}}$ is the set of triangle visible in $c_{key}$ with $T_{c_{key}} \in T_z$.
$T_z$ is a fixed fraction $z$ of all the available triangle set $T_{all}$ (i.e. $\lvert T_z\rvert = \lvert T_{all}\rvert / z $).

Of all drawn combinations of matching partners, we now select the one that maximizes this combined fulfillment. 
The combination of key view and selected combination, we will further call view cluster.
We execute this step for all available images.
%
%

\subsection{Next-Best View Ranking}

In general, image acquisition for photogrammetric reconstruction is done with a lot of redundancy to
ensure that the whole scene is sufficiently covered.
However, a drawback of this acquisition technique is that some images do not contain any additional 
information compared to their neighbors.
Finding the best and minimal subset for a sufficiently complete reconstruction is a very hard task.
In fact, subproblems of this task are known to be NP-hard.
E.g. if an oracle told us that the minimal number of key views for sufficiently covering the scene is $n$,
we would still have to solve the NP-hard maximum coverage problem~\citep{feige98}.
While solving this problem optimally is computationally intractable (unless P = NP),
there exist approximation approaches with theoretic quality guarantees for the found solution.
In particular, if the objective function is a nonnegative monotone submodular function, then a  
greedy algorithm is guaranteed to select $n$ views such that
the reached objective value is within 63\% of the optimal objective value for the same amount of views~\citep{nemhauser1978analysis}.
This fact (and the fact that $n$ is not easily estimated) motivated us to 
solve this problem with a greedy algorithm and a nonnegative monotone submodular objective function.

\paragraph{Objective function}
Now let us formalize the objective function which we aim to maximize, i.e. the overall fulfillment, as:
\begin{equation}
 f_o(V,T) = \frac{1}{\lvert T \rvert} \sum_{t \in T}  \max_{v \in V} f(t,v) ,
 \label{eq:nbv_fulfillment}
\end{equation}
where $T$ is the set of all mesh triangles,
$\lvert T \rvert$ is the size of this set,
$t$ is one of these triangles, $V$ is the set of view clusters and
$v$ is one of these view clusters.
$f(t,v)$ is the triangle fulfillment as defined in Equation~\ref{eq:triangle_fulfillment} with $v = \{c_{key},c^k\}$.
Note that maximizing this objective function for any fixed number of view clusters
results in an NP-hard problem, however, 
the quality guarantees of~\cite{nemhauser1978analysis} for a greedy algorithm still apply.



\paragraph{Algorithm}
We formulate our ranking procedure as a greedy algorithm.
In each iteration, we select the next-best view cluster, i.e. the view cluster with the highest fulfillment gain.
We define this gain for a view cluster $v_i$ as: 
\begin{equation}
 g(v_i,V',T) = f_o(\{V' \cup v_i\},T) - f_o(V',T),
\end{equation}
where $V'$ is the set of already selected view clusters.
For an efficient solution to this problem,
we propose the following algorithm.

Our algorithm first starts by estimating the visibility between each view/camera and the 
triangle mesh.
This visibility information is stored in each triangle (i.e. each triangle knows which 
camera has a direct line of sight to it).
Based on this information, we precompute the fulfillments $f(t,v)$ for all triangles $t \in T$ 
with sufficient camera views ($\geq x$, related to the coverage fulfillment $f_{cov}$)
and all view clusters.
This fulfillment $f(t,v)$ is then stored within a map in each triangle $t$.

For each view cluster $v$, we now compute the initial fulfillment gain $g(v,V',T)$ with $V' = \{\}$ (i.e. an empty set of selected view clusters).
Each pair of view cluster and gain values $\{v,g(v,V',T)\}$ is inserted into a priority queue, 
where now each view cluster is ranked by its fulfillment gain.
While building up this data structure has a complexity of $\mathcal{O}(n\log{}n)$ with $n = \lvert V \rvert$,
we can now use lazy updates in each iteration and avoid a significant number of unnecessary computations.

In each iteration, we now select and pop the top element of the queue, i.e. the element with the highest fulfillment gain,
and add the selected view cluster $v_s$ to the set of selected view clusters $V'$.
Then we update the current fulfillment for each triangle $t$ that is observed by $v_s$.
This current fulfillment is stored inside the triangle $t$ and computed as $f'(t,V') = \max_{v \in V'} f(t,v)$.
Based on this information, we now lazily update all elements of the priority queue.
This means that we select and pop the top element of the queue and update the fulfillment gain of the 
corresponding view cluster $v_u$. We store all such temporarily removed view clusters $V_u$ in a separate data structure and
keep track of the maximum fulfillment gain $g_{max}$ over all these values.
We can stop the update procedure if $g_{max} \geq g(v_{top},V',T)$, where $v_{top}$ is the current top element of the priority queue.
Then all temporarily removed view clusters $V_u$ are reinserted into the priority queue.
Now the next iteration can start.
We terminate if $g(v_{top},V',T) = 0$ or the priority queue is empty.

Note that all computational components (the visibility estimation, the confidence prediction and the fulfillment computation)
are computed only once at the beginning of algorithm and the execution of the actual MVS algorithm is avoided altogether.
This makes the ranking procedure very light weight in terms of computation time.
The final output of this ranking procedure is an ordered list of view clusters (key views with matching partners) together 
with the estimated fulfillment up to this point.

\section{Experiments}

In our experiments, we use two different environments.
The first environment contains cultural heritage sites 
in the valley of Valcamonica, Italy.
The cultural heritage sites consist of hundreds of open rock surfaces covered with prehistoric rock art.
The sites all look very similar to each other and contain a limited amount of structures;
i.e. the sites themselves are open rock surfaces are surrounded by lawn, bushes, wooden man-made structures (e.g.  hand rails) and 
trees.
The high amount of vegetation makes 3D reconstruction in this environment very hard.
Thus, this environment represents the ideal use case for our machine learning approach,
as it is a hard and reoccurring task.
For our experiments, we use the confidence predictor of~\cite{mostegel16b}, which was specially trained for this kind of environment.
For evaluation, we then use a test site which was not included in the training.
This allows us to demonstrate the full potential of our approach in the same environment in which is was trained.

The second environment, we use for evaluation, is a suburban setting of single family houses.
We will use this scenario only for evaluation and not for training.
With this setting we evaluate the domain generalization properties of our confidence predictor.
I.e. how does the predictor react if it is confronted with structures it has 
never seen before? 

Using these two datasets, we structure our experiments in three main parts.
First, we evaluate only the matching partner selection in Section~\ref{ss:matching} and 
then only the ranking performance in Section~\ref{ss:ranking}.
From both of these experiments,
we then take the best performing baseline approach and evaluate this combination against our full approach in Section~\ref{ss:full}.

\subsection{Evaluation Details}
\label{ss:eval_details}

For all our experiments, we use the same confidence predictor.
This confidence predictor was trained for the environment of Valcamonica on
5000 images of 8 different sites~\citep{mostegel16b} with SURE~\citep{rothermel12} as MVS algorithm.
Note that this dataset does not have any overlap with the two datasets.
We selected SURE as our main MVS algorithm as it a widely accepted photogrammetric software,
which yields high quality results.

For the matching partner selection, we use the $n = 22$ most connected images, draw $y = 100$ combinations
and set the triangle fraction factor to $z = 10$.

For our fulfillment functions, we set the necessary number of cameras to $x = 3$ in line with the default parameter of SURE~\citep{rothermel12} (i.e. 2 matching partners).
We set the desired ground sampling distance $g_d$ and desired accuracy $a_d$ to 1cm and the weighting parameter $\alpha$ to 0.5.

\begin{figure}[t]
  \centering
  \subfigure[Valley Dataset]
    {
                \includegraphics[width=0.508\columnwidth]{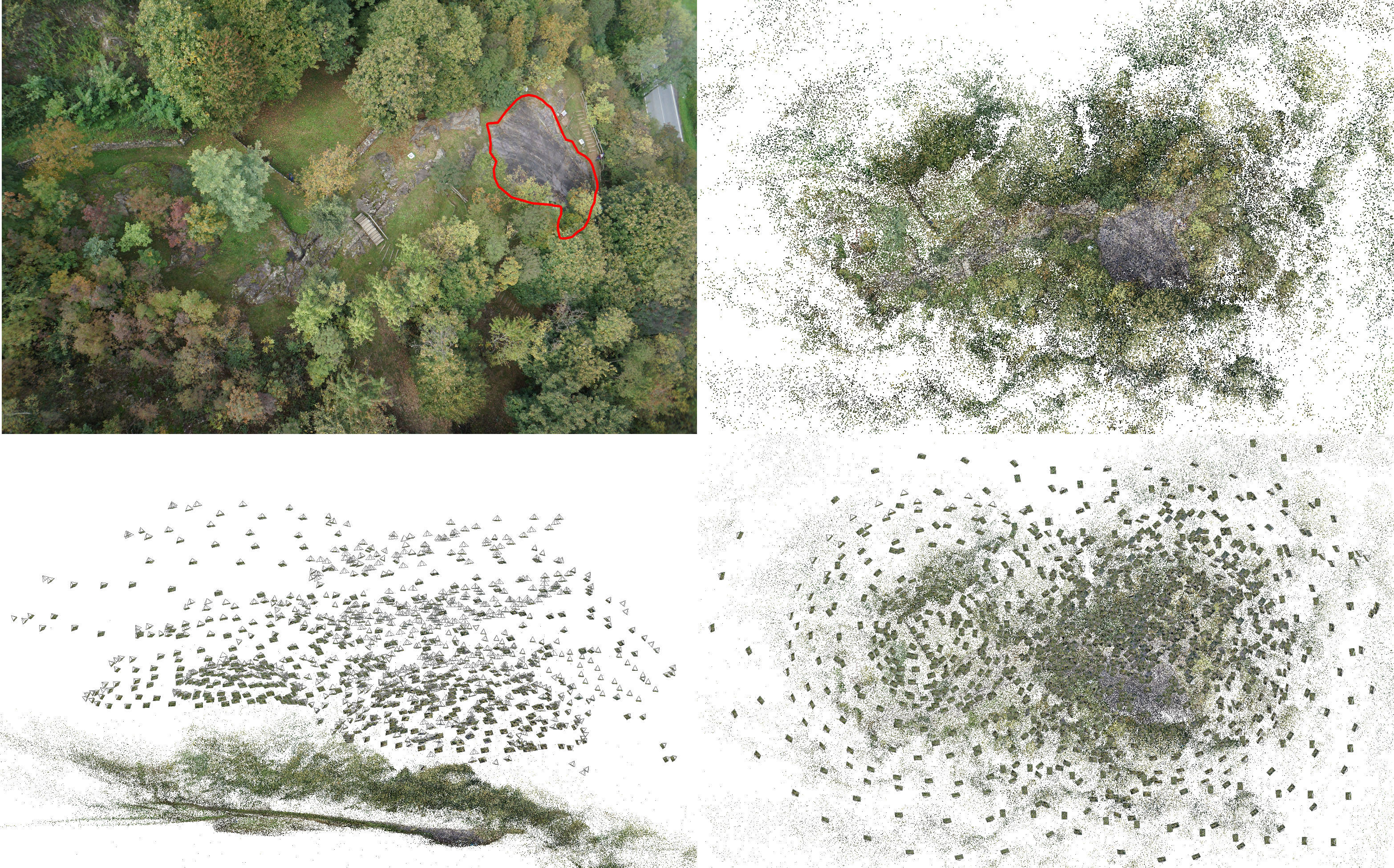} 
}
\quad
\subfigure[House Dataset]
    {
                \includegraphics[width=0.4\columnwidth]{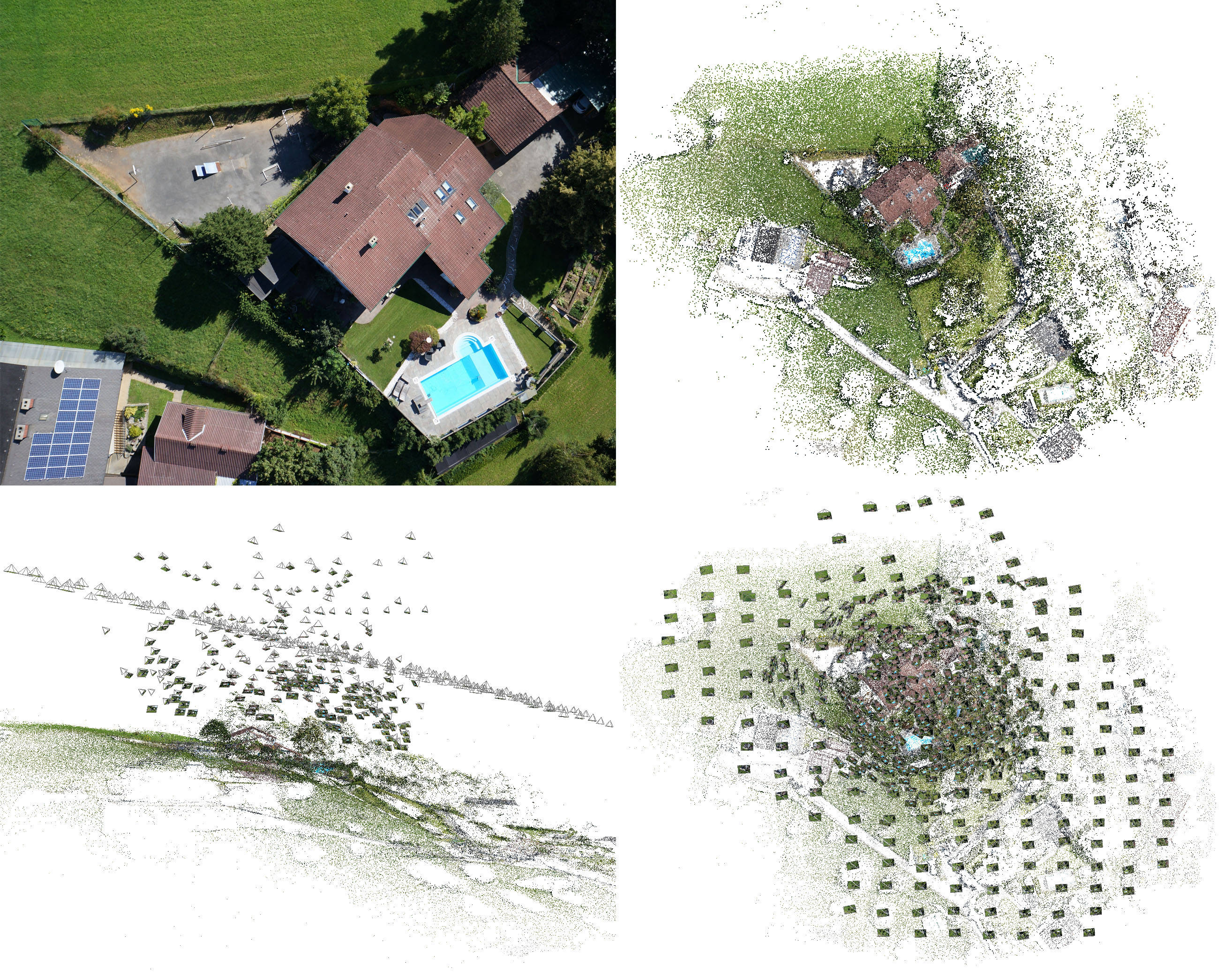} 
}
  \caption{Valley and House Dataset. Top left: One image of the dataset shows an overview of the scene. Top right: A nadir view of the sparse point cloud.
  Bottom right: The camera poses with the sparse point cloud from the same nadir view point. Bottom left: A side view of the camera poses with the sparse point cloud.
  For the Valley Dataset, ground truth of the rock formation Seradina 12C (marked in red) was acquired with a laser scanner.}
  \label{fig:valcamonica_1k}
\end{figure}

\paragraph{Valley Dataset}
The dataset consists of 1236 registered images of a complex scene in the valley of Valcamonica (Figure~\ref{fig:valcamonica_1k}).
The images were acquired on 3 consecutive days in regular patterns (i.e. grids and domes) and 
in semi-structured ways using autonomous image acquisition~\citep{mostegel16b}.
The grid was acquired with 80\% overlap and 8mm GSD.
The domes were centered on two separate locations and the GSD was varied approximately from 8mm to 16mm.
For the semi-structured autonomous image acquisition~\citep{mostegel16b}, 
the quality parameters were set to 8mm ground sampling distance and accuracy.
Note that this dataset has no overlap with the dataset used for training the confidence predictor.
The images were processed with structure-from-motion pipeline of~\citep{rumpler17} and 
geo-referenced using the same fiducial markers as in~\citep{mostegel16b}.
This resulted in a sparse point cloud of approximately 480k points,
where nearly all cameras contain 3D points with 100+ connections.

\paragraph{House Dataset}
The House Dataset contains 485 registered images of a detached house with 3000 $m^2$ garden.
The images were taken in a regular grid and several iterations of our autonomous image acquisition approach~\citep{mostegel16b}.
The grid was acquired with 80\% overlap and 1cm GSD, and for our approach with 1cm for ground resolution and accuracy.
While some scene structures are similar to the Valley environment (grass and trees),
most structures have never been seen in training (roofs, house walls, swimming pool, cars, etc.).

\paragraph{Computation Time}
We ran all our experiments on the same desktop computer with a Intel Core i7-4771 CPU,
32GB of RAM and GeForce GTX 770.
We use the linux version of SURE~\citep{rothermel12}, which only uses the CPU.
With this version of SURE the average depth map computation time per matching partner on our datasets is 46.6s.
In our approach, the mesh computation and the visibility casting are negligible compared to the more computationally intense parts of our approach
(both parts together are finished in less than two minutes for both datasets).
Our confidence prediction takes a constant time of 1s per image.
In Table~\ref{tab:timings}, we show the time consumption for the matching partner selection and 
the next-best view ranking for each dataset and number of matching partners separately.
Note that our prioritization approach is very light weight and only requires 2\%-6\% of the time the 
matching procedure of SURE~\citep{rothermel12} itself.
This means that if we reduce the number of required key views to 5\%, we effectively 
reduce the overall run-time to 7\%-11\%, which speeds up the full MVS reconstruction process by a full order of magnitude.

\begin{table}
\centering
 \begin{tabular}[b]{|c||c|c|c|c|}
  \hline
   Num Matching Partners  &  2 &  3  &   5 & 11 \\\hline 
    \multicolumn{5}{|c|}{Matching Partner Selection [s]}  \\\hline\hline 
    Valley Dataset &  0.7 & 1.0 & 1.8 & 23.8\\\hline  
    House  Dataset &  0.7 & 1.0 & 1.8 & 18.2\\\hline  
    \multicolumn{5}{|c|}{Next-Best View Ranking [s]}  \\\hline\hline 
    Valley Dataset &  3.5 & 3.6 & 4.0 & 10.8\\\hline  
    House  Dataset &  2.8 & 3.1 & 3.4 & 8.5\\\hline  
    \multicolumn{5}{|c|}{Overall Average [s]}  \\\hline\hline 
    (Valley+House)/2 &  3.9 & 4.4 & 5.5 & 30.7\\\hline  
    \multicolumn{5}{|c|}{Prioritization Time / Matching Time [\%]}  \\\hline\hline 
    (Valley+House)/2 &  4.13\% & 3.11\% & 2.36\% & 5.98\%\\\hline  
\end{tabular}
\caption{We show the average time consumption of the two main parts of our approach (i.e. Matching Partner Selection and Next-Best View Ranking)
in seconds per key view. All initialization steps are included in the timings. The last two row show the overall average
and the relative time consumption between our prioritization method and the matching time of SURE~\citep{rothermel12}.
 }
\label{tab:timings}
\end{table}

\subsection{Selecting the k-best Matching Partners}
\label{ss:matching}
\begin{figure}[t]
       \vspace{-50pt}
  \centering
\subfigure[Valley Laser GT]
    {
                \includegraphics[width=0.33\columnwidth]{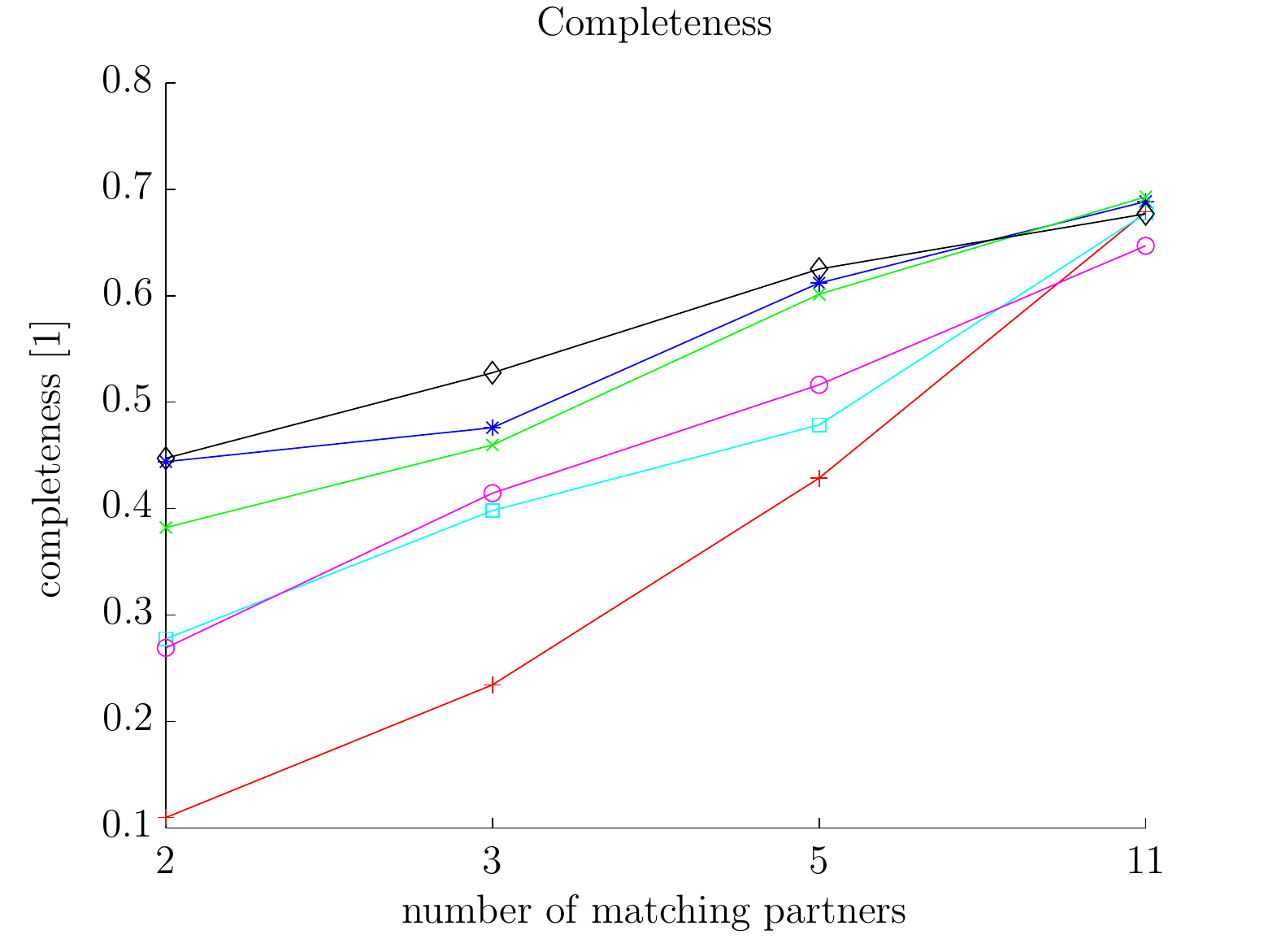} 
}
\quad
\hspace{-38pt} 
 \subfigure[Valley 22 MP]
    {
                \includegraphics[width=0.33\columnwidth]{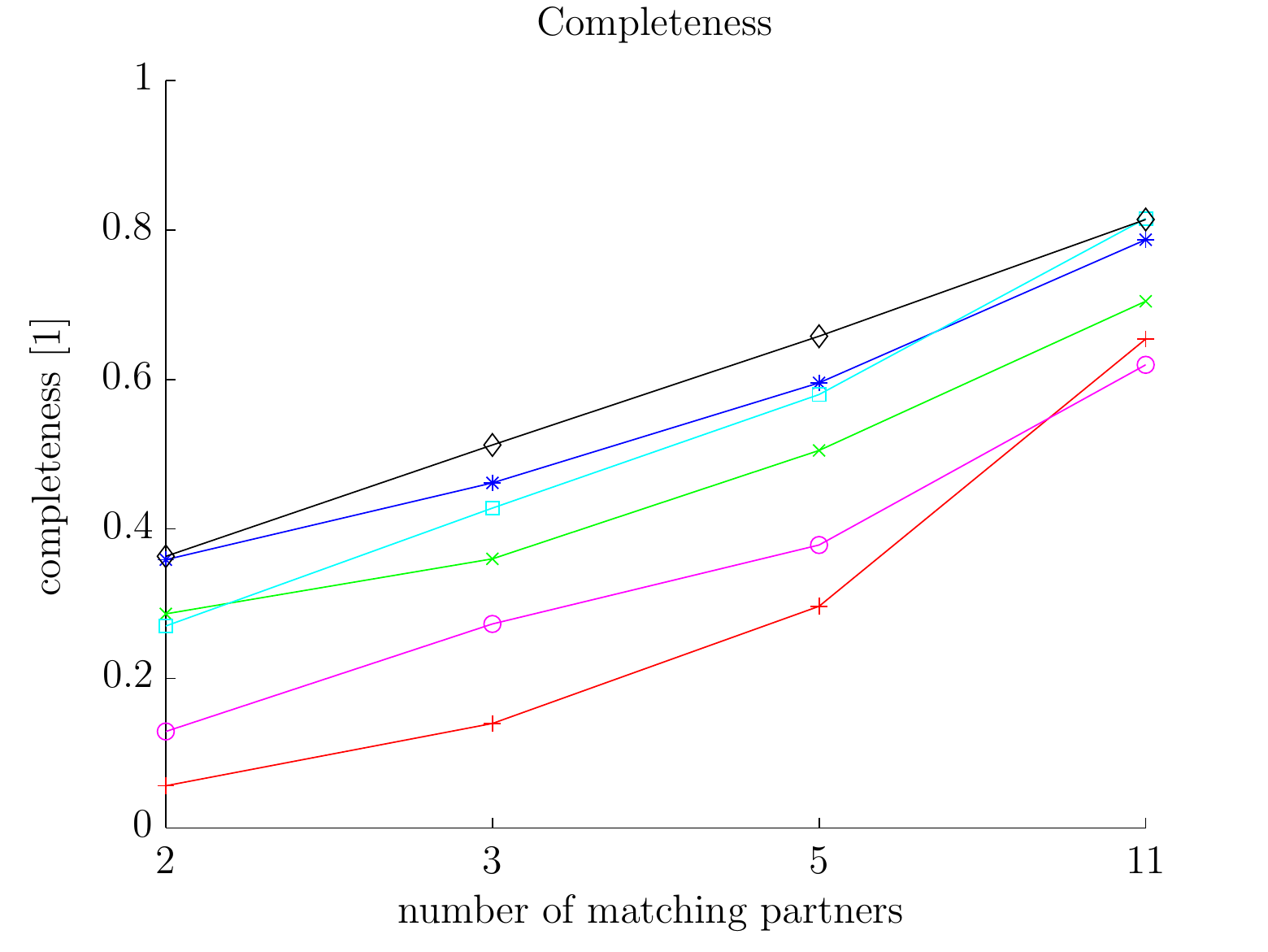} 
}
\quad
\hspace{-38pt} 
\subfigure[House 22 MP]
    {
                \includegraphics[width=0.33\columnwidth]{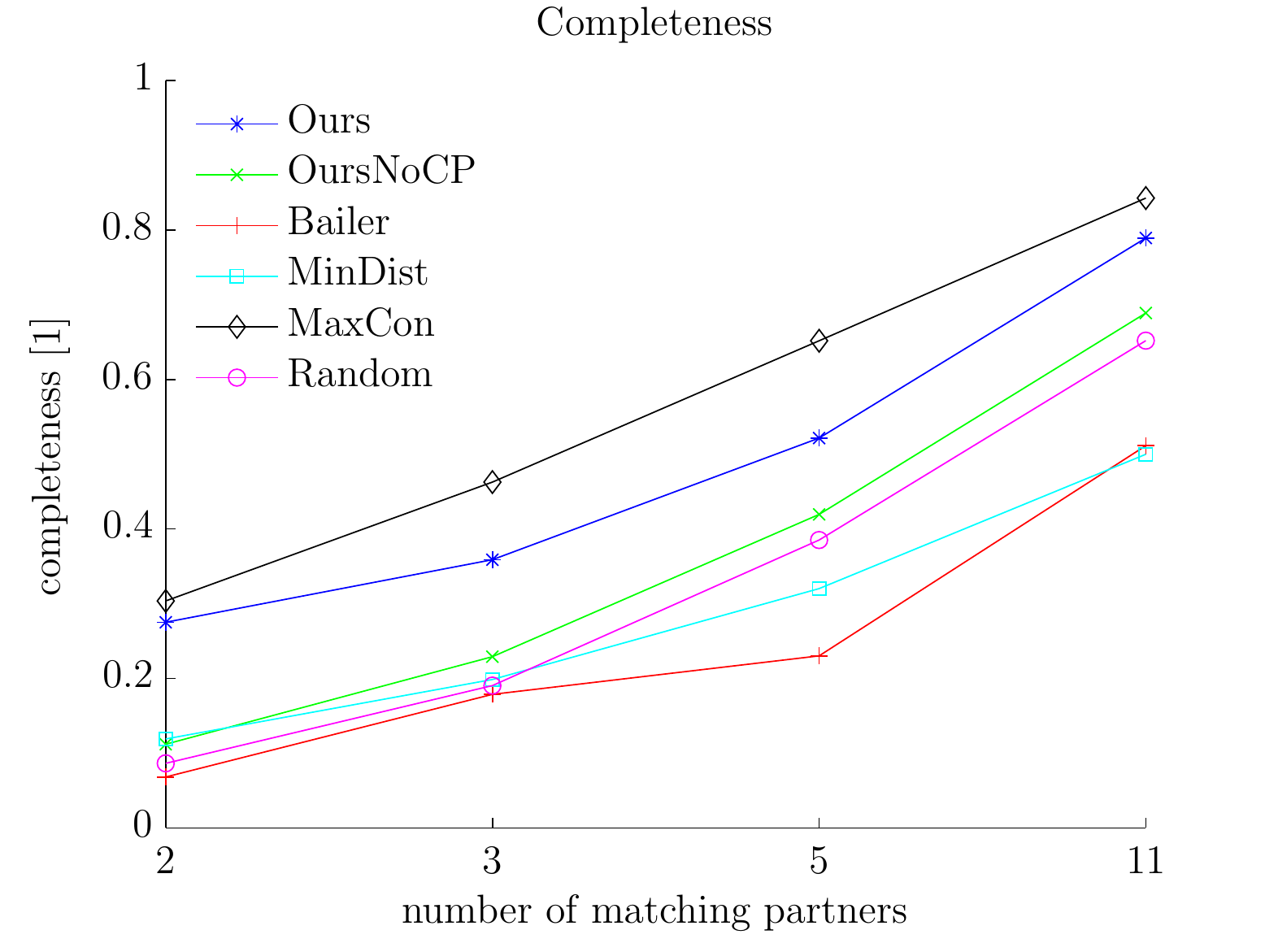} 
}
\quad
\subfigure[Valley Laser GT]
    {
                \includegraphics[width=0.33\columnwidth]{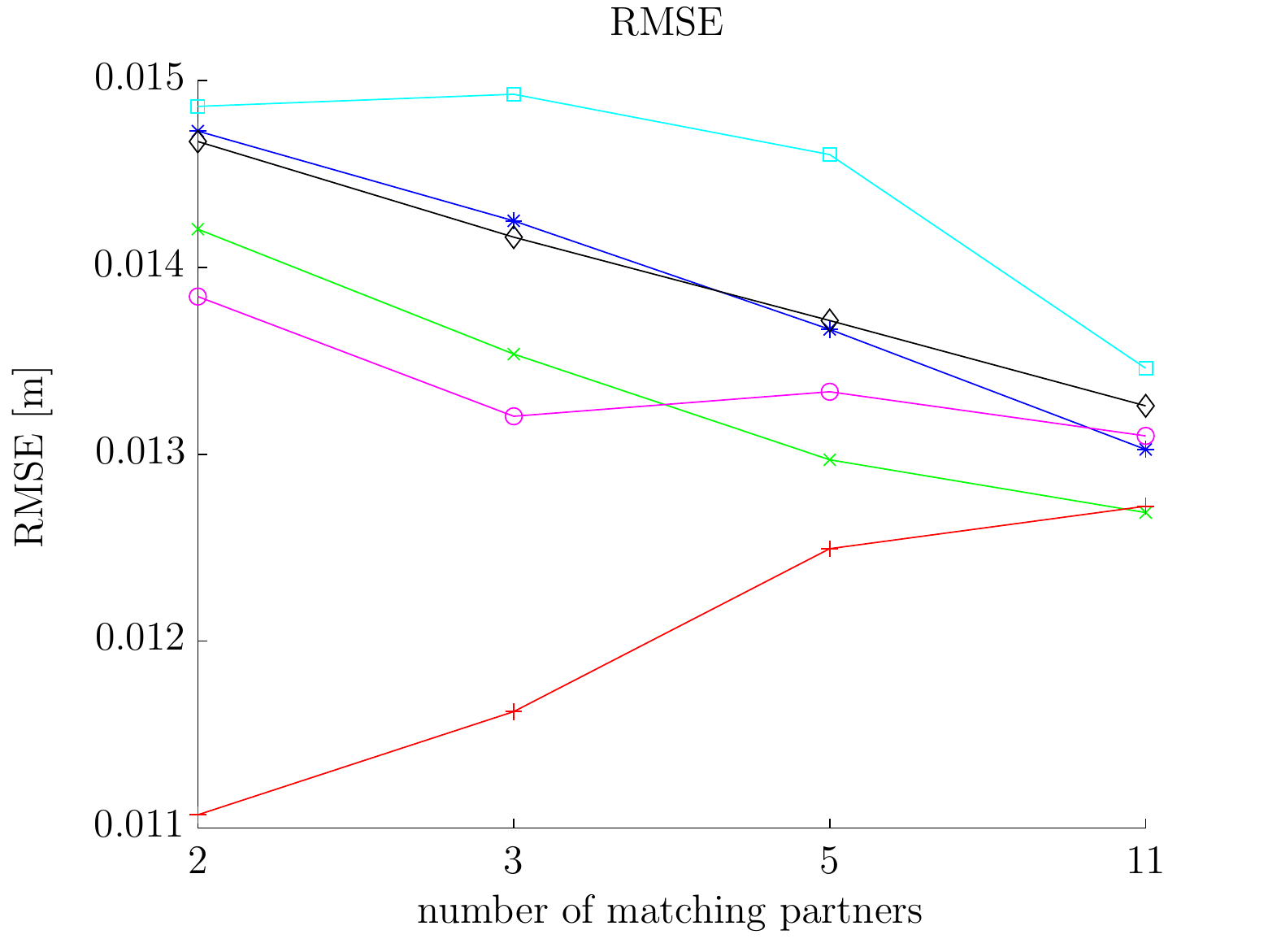} 
}
\quad
\hspace{-38pt} 
 \subfigure[Valley 22 MP]
    {
                \includegraphics[width=0.33\columnwidth]{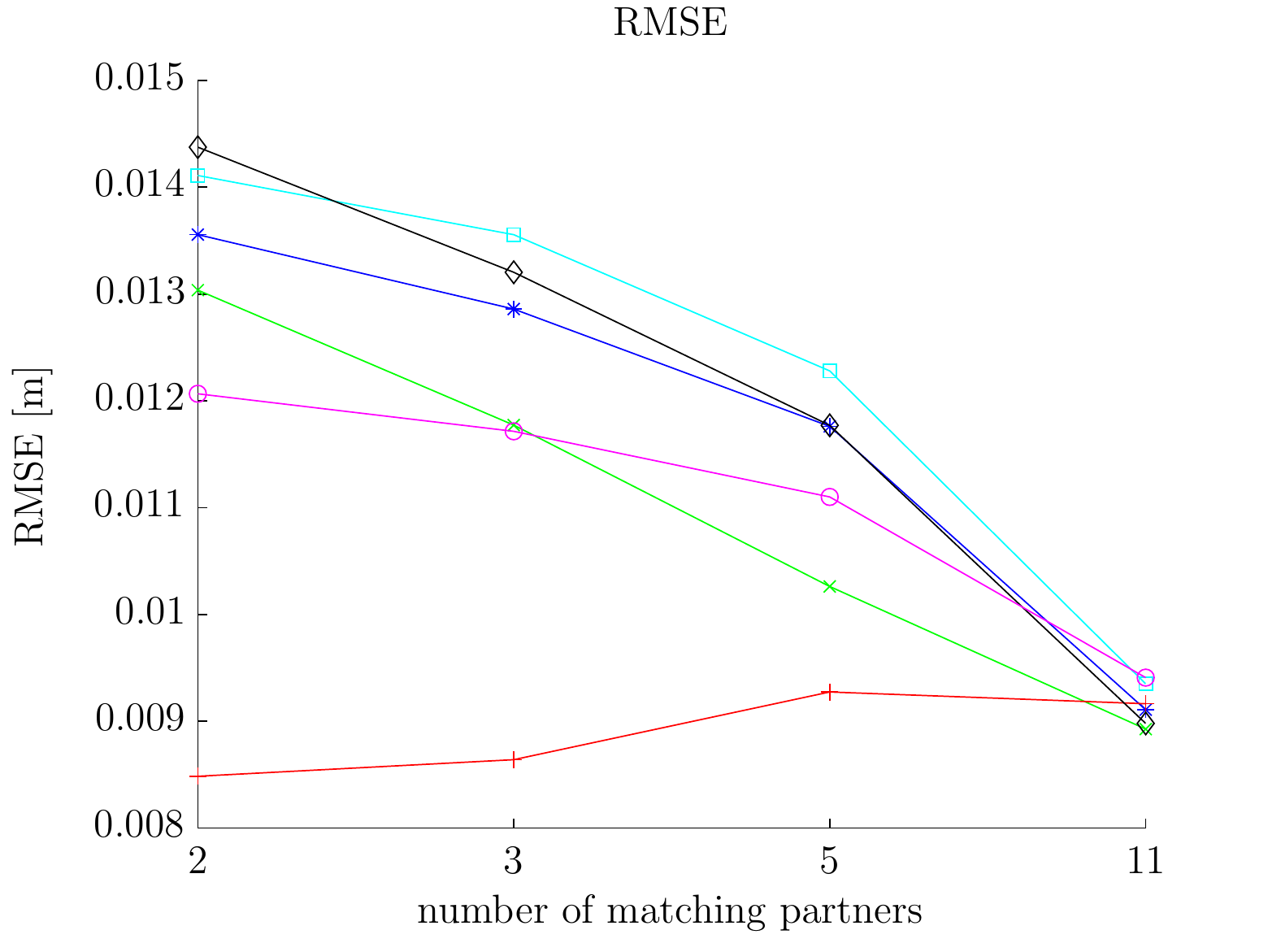} 
}
\quad
\hspace{-38pt} 
\subfigure[House 22 MP]
    {
                \includegraphics[width=0.33\columnwidth]{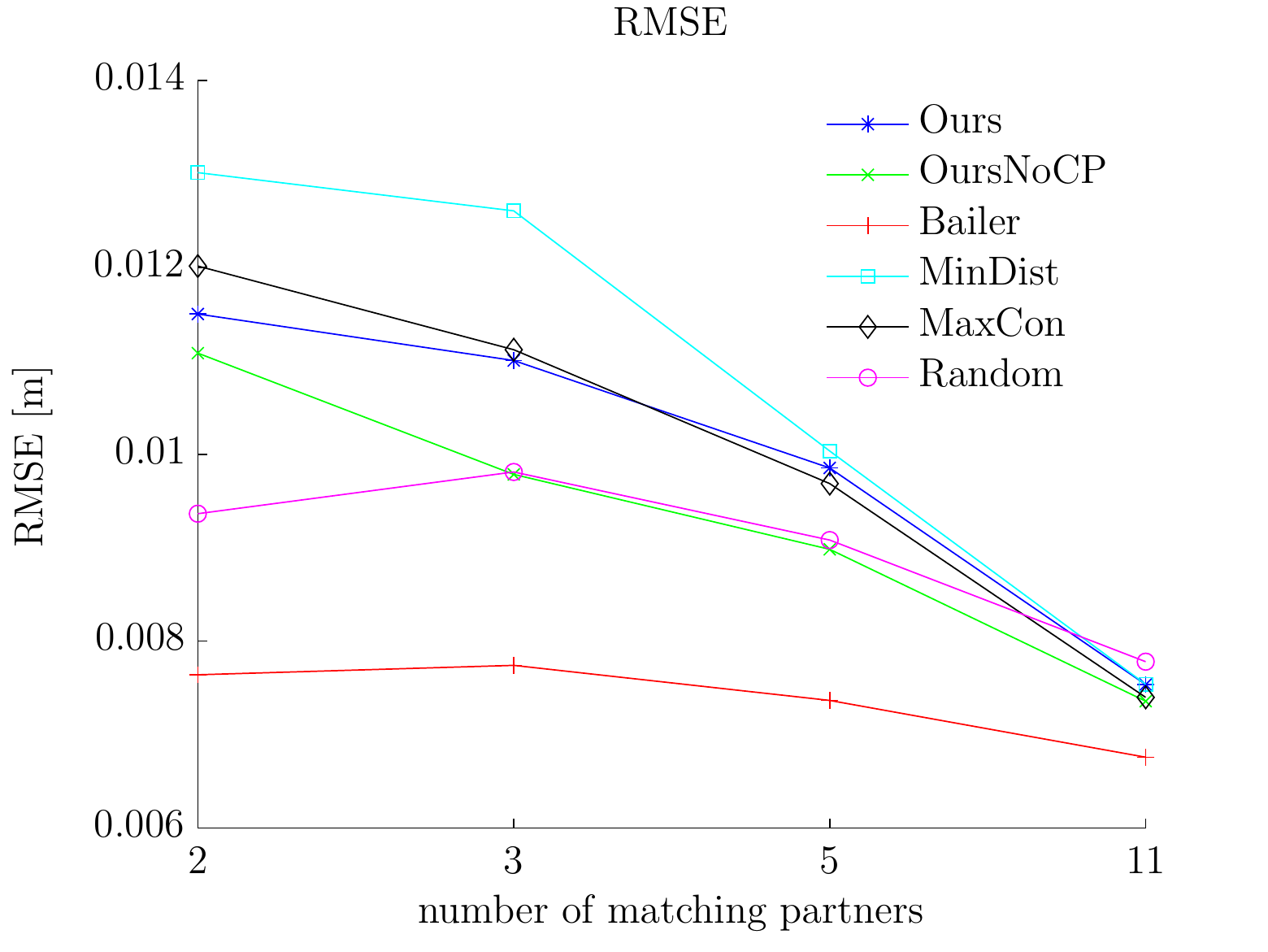} 
}

\quad
\subfigure[Valley Laser GT]
    {
                \includegraphics[width=0.33\columnwidth]{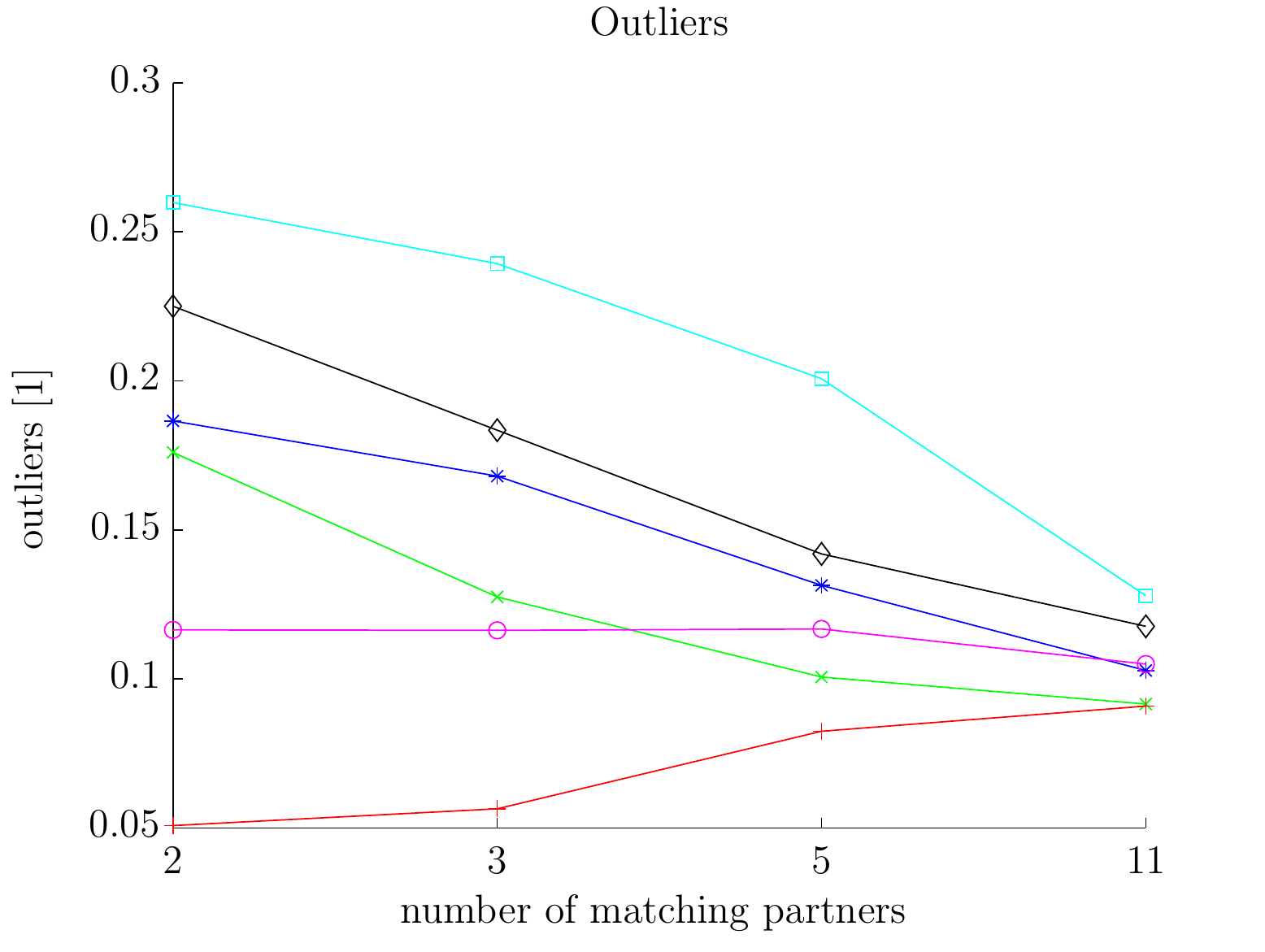} 
}
\quad
\hspace{-38pt} 
 \subfigure[Valley 22 MP]
    {
                \includegraphics[width=0.33\columnwidth]{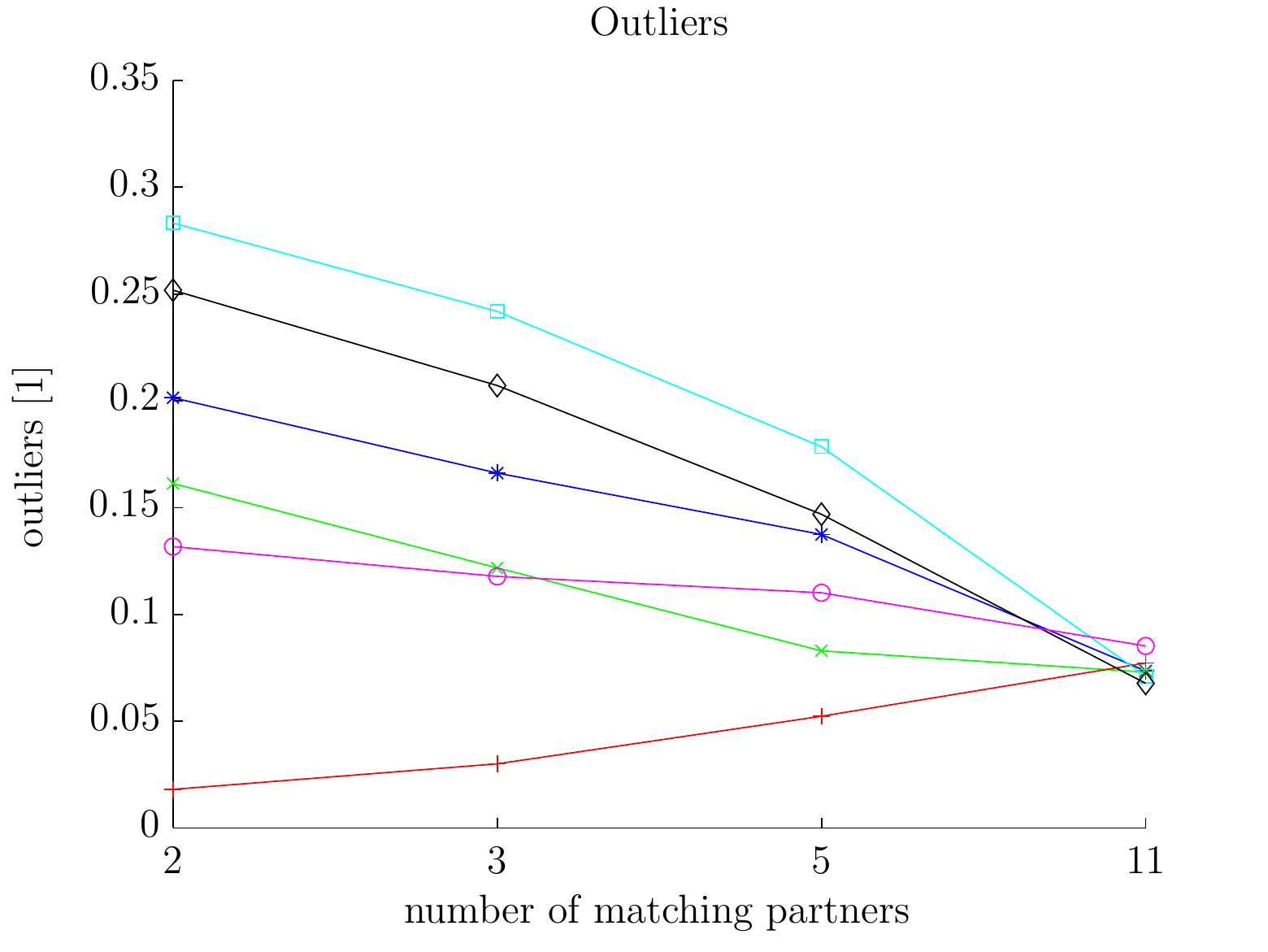} 
}
\quad
\hspace{-38pt} 
\subfigure[House 22 MP]
    {
                \includegraphics[width=0.33\columnwidth]{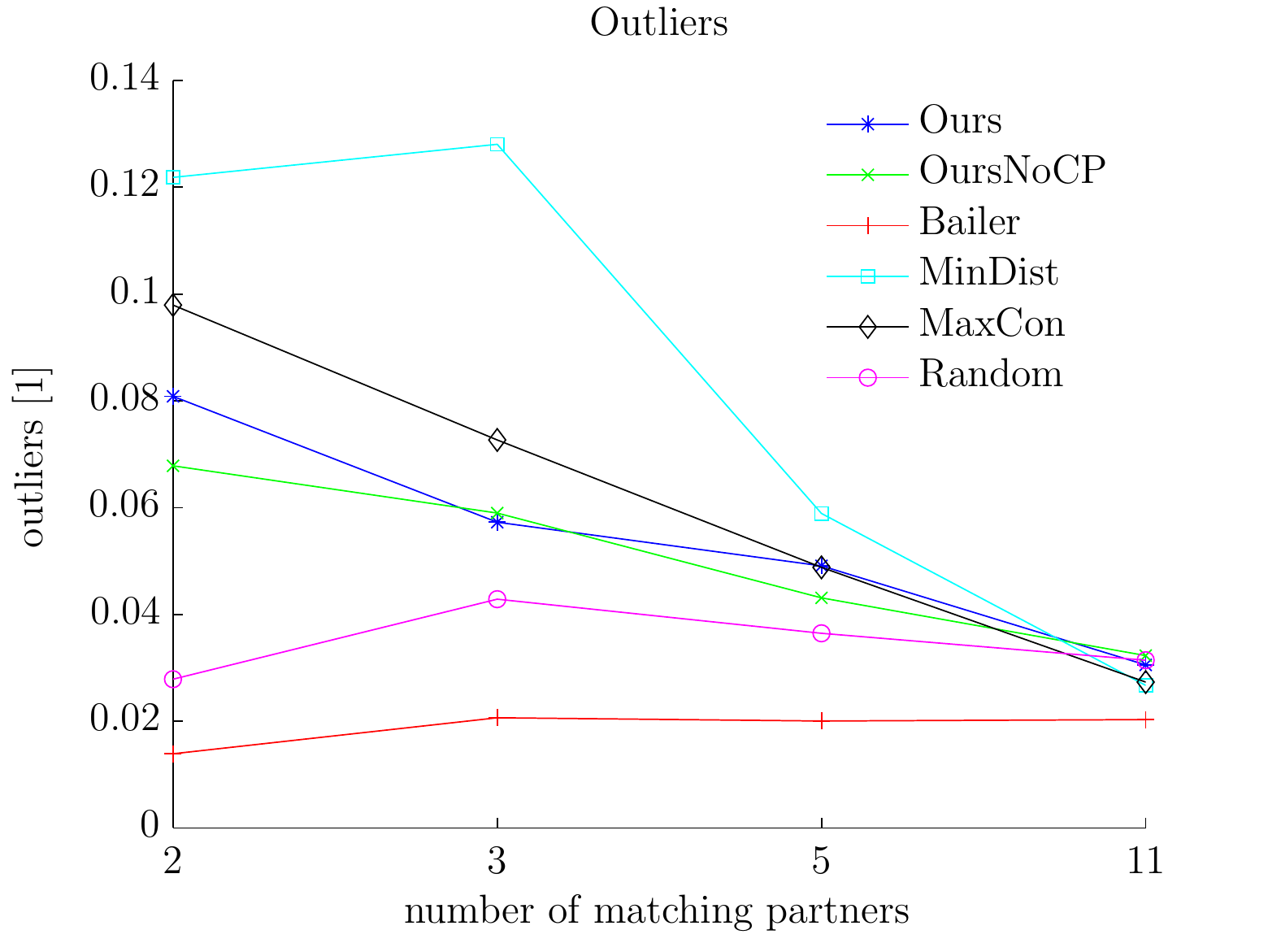} 
}
    \caption{Matching partner selection experiment on the {\bf Valley} and {\bf House} Datasets. We show completeness, RMSE (Root-Mean-Square Error) and outlier ratio
    compared to the output with the 22 most connected matching partners (22 MP) and
    the laser ground truth of a rock formation (Laser GT). 
    }
  \label{fig:mp_valley}
\end{figure}

In this experiment, we evaluate the first subtask of our view cluster prioritization, i.e. the matching partner selection.
In this subtask, the algorithm is given a fixed set of key frames and for each key frame the $k$-best matching partners should be selected.
Thus we evaluate three factors in this experiment, i.e. completeness, outlier percentage and RMSE (root mean square error).
However, this evaluation requires ground truth.
For the Valley Dataset, we have a traditional ground truth only for a small part of the scene,
i.e. a rock formation called  Seradina Rock 12C (Figure~\ref{fig:valcamonica_1k}).
This ground truth was captured with a Riegl VZ-400 Laser Scanner and has an accuracy of 5mm.
While we also use this ground truth in our evaluation (denoted as "Laser GT"), it is limited to a mostly flat rock surface.

In order to evaluate the three quality factors also for the other parts of the scene (including the vegetation),
we have to use a more unconventional approach.
For this purpose, let us first formally define the task of the matching partner selection.
In the task of matching partner selection, an algorithm shall select a subset $M_k$ of $k$ matching partners (images) 
for a specific MVS algorithm such that the resulting depth map 
maximally resembles the depth map produced with the same algorithm and a much larger set of matching partners $M_{large}$ with $M_k \subset M_{large}$
and $\lvert M_k \rvert \ll \lvert M_{large} \rvert$ .
With this definition, we can use the depth map produced with $M_{large}$  as "ground truth"
for assessing the quality parameters.
We will further refer to this ground truth as "22 MP" as we use the most connected 22 matching partners.

Using these two kinds of ground truth, we evaluate the three quality parameters in the following manner.
First, we detect outliers with respect to the ground truth and the desired accuracy.
I.e. we classify all depth estimates which are more than 3 times the desired accuracy (i.e. $> 3 \cdot 1$cm) as outliers.
Note that for the traditional ground truth, we additionally exclude occlusions that are more than 24cm from the rock surface from this evaluation.
Then we evaluate the completeness and RMSE with respect to the ground truth
using only the valid measurements (without outliers and occluded values).

From the large set of possible approaches for matching partner selection (see Section~\ref{ss:ic_related_work}),
we select the most relevant approaches for the given algorithm and application of photogrammetric reconstruction.
Thus, we evaluate our approach against two photogrammetric standard approaches, which are both implemented by SURE~\citep{rothermel12},
i.e. the $k$ closest images (MinDist), the $k$ most connected images (MaxCon).
Further, we select $k$ random images as proposed by~\citep{galliani15} (Rand) and $k$ images with the carefully hand-crafted approach 
of~\citep{bailer12}.
Finally, we run our approach once with confidence prediction (Ours) and once without prediction (OursNoCP), i.e. $f_{conf} = 1$.
For each key frame, all approaches have the task of selecting $k = {2,3,5,11}$ out of the 22 most-connected matching partners.

For computational reasons, we do not computed all possible combination on all images of the datasets, but use a representative subset.
For the \emph{22 MP} evaluation, we will use 100 randomly selected key views for evaluation.
For the \emph{Laser GT} evaluation (of Rock 12C), we select the 50 views (of the 100) in which Seradina 12C is most prominently visible (i.e. covers the largest area in pixel).

\begin{figure}[t]
   \vspace{-20pt}
  \centering

  \includegraphics[width=0.78\columnwidth]{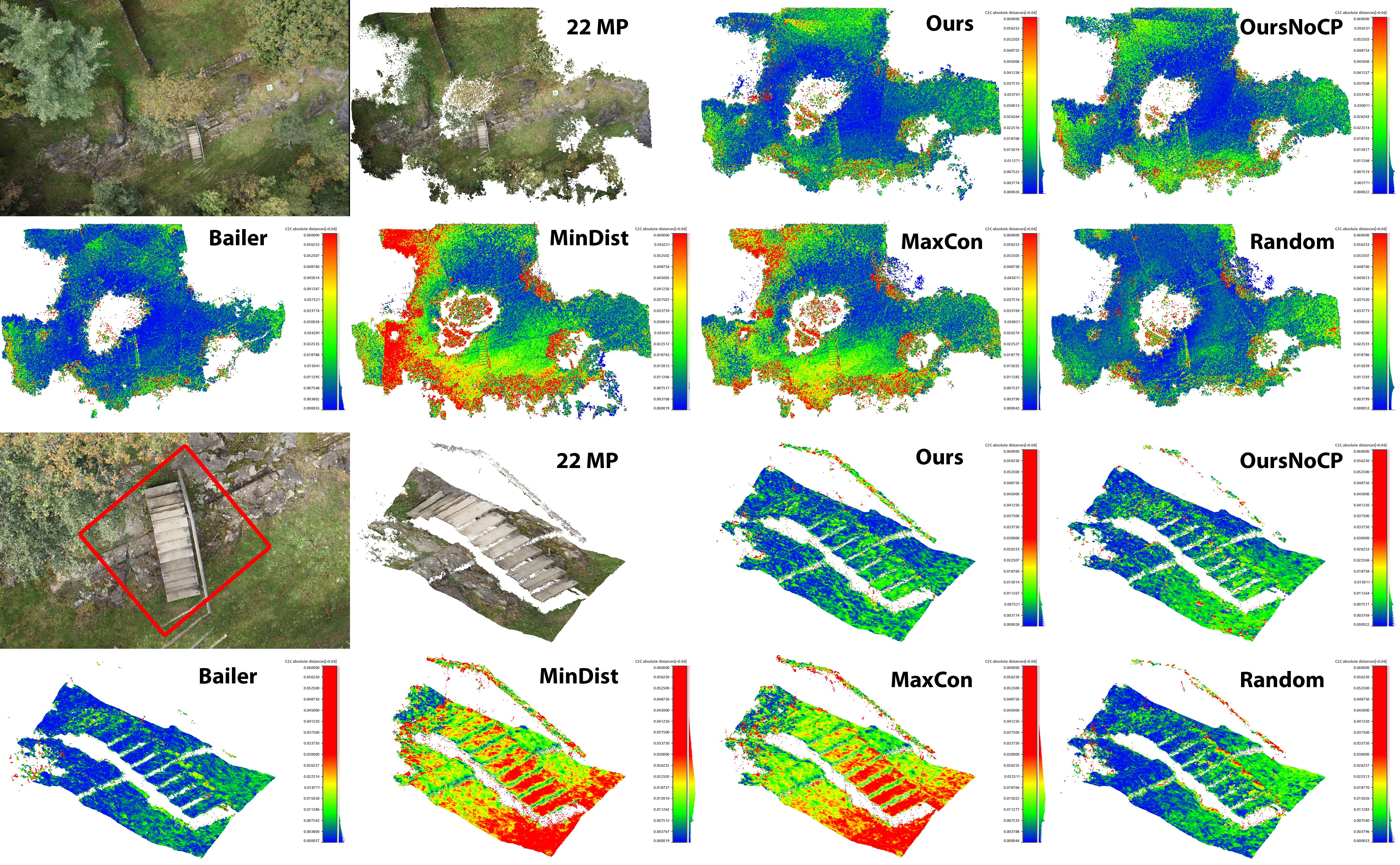} 
  \caption{Impact of matching partner selection (selecting 5 out of 22) on the {\bf Valley Dataset}. All shown reconstructions are computed with the same image as key view (top left).
  The top two row show the whole point clouds where the color encodes the distance to the reference (22MP);
  from no error (blue) to \emph{6cm} error (red). The bottom row shows a cutout of the same reconstruction (here red indicates an error larger than \emph{3cm}). 
  Notice that our approach leads to a reconstruction with low error that still preserves fine details such as the hand rail.
  Other approaches have either a low error but also low completeness (Bailer) or have a high completeness but also a high error (MinDist and MaxCon). 
  }
    \vspace{-15pt}
  \label{fig:mp_comparison}
\end{figure}

\paragraph{Results}
In Figure~\ref{fig:mp_valley}, we show the averaged results 
in completeness, the root mean square error (RMSE) and 
the ratio of outliers.

Now let us first compare the results of \emph{Valley Laser GT} and \emph{Valley 22 MP}
to establish the relation between traditional ground truth and the ground truth with a large number of matching partners.
In this regard, the number of outliers seems to be most important.
If we compare sub-figures (g) and (h), we can observe that 
the relative ordering between the approaches stays the same for 2,3 and 5 matching partners.
From this we conclude that our approximative ground truth seems to work reasonably up to 5 matching partners.

Now let us compare the individual approaches on the Valley Dataset.
As expected, for all approaches more matching partners lead to higher completeness.
Similarly, more matching partners lead to a lower outlier ratio for most approaches.
Only for Bailer, the number of outliers seems to increase (we think this is due to the very low completeness).
For all approaches, the RMSE and the outlier ratio seem to be highly correlated.

In general, we can observe a tradeoff between completeness and accuracy (outliers and RMSE).
None of the evaluated approaches leads in both completeness and accuracy.
Some approaches are rather tuned for completeness (MaxCon and Ours), while
others more for accuracy (Bailer and OursNoCP).
In contrast to only flat surfaces (Laser GT),
we can see a clear gap between the two versions of our approach on the complete scene (22 MP).
Further, 
we can see that our approach seems to be tuned very much towards completeness (very close to MaxCon),
but at the same time shows a significantly lower outlier ratio than MaxCon (especially for a low number of matching partners).

If we now take a look at the results of the House Dataset,
the relative ordering between MaxCon, Ours and OursNoCP is very similar.
Ours stays between the other two approaches (for completeness and accuracy),
but exhibits a much higher completeness than OursNoCP.

Overall we conclude that our approach with learning leads to a significantly higher completeness than our 
approach without learning and at the same time keeps the outlier ratio significantly lower than 
approaches of similar completeness (especially for a low number of matching partners).
This can also be observed in the example shown in Figure~\ref{fig:mp_comparison}.

\subsection{Ranking Only}
\label{ss:ranking}
In the previous experiment, we fixed the set of key views and evaluated the matching partner selection.
In this experiment, we fix the matching partner selection and evaluate the cluster ranking performance.
For each key view, we run our full approach with 5 matching partners.
This leads to one view cluster per key view.

The task in this experiment is to maximize the \emph{real} fulfillment with as few view clusters as possible.
We evaluate the real fulfillment analogue to Equation~\ref{eq:nbv_fulfillment}, with the difference that all estimated and predicted values are replaced
by measurements as follows.
For evaluating the coverage, we use the projected depth map together
with the set of successful matching partners reported by SURE.
For judging whether a measurement represents a valid measurement of a triangle we use the same criterion as in the last experiment.
For evaluating the theoretical uncertainty (Equation~\ref{eq:aia_func}), we use the set of reported matching partners by SURE
and check for self occlusions for each triangle.
Finally, we replace $f_{conf}$ with the actual coverage of the triangle.
This means for a given triangle, we project this triangle into the key view and analyze 
the percentage of pixels with a valid measurement within the 2D projection of the triangle.

\begin{figure}[t]
  \centering
 \subfigure[Valley]
    {
                \includegraphics[width=0.5\columnwidth]{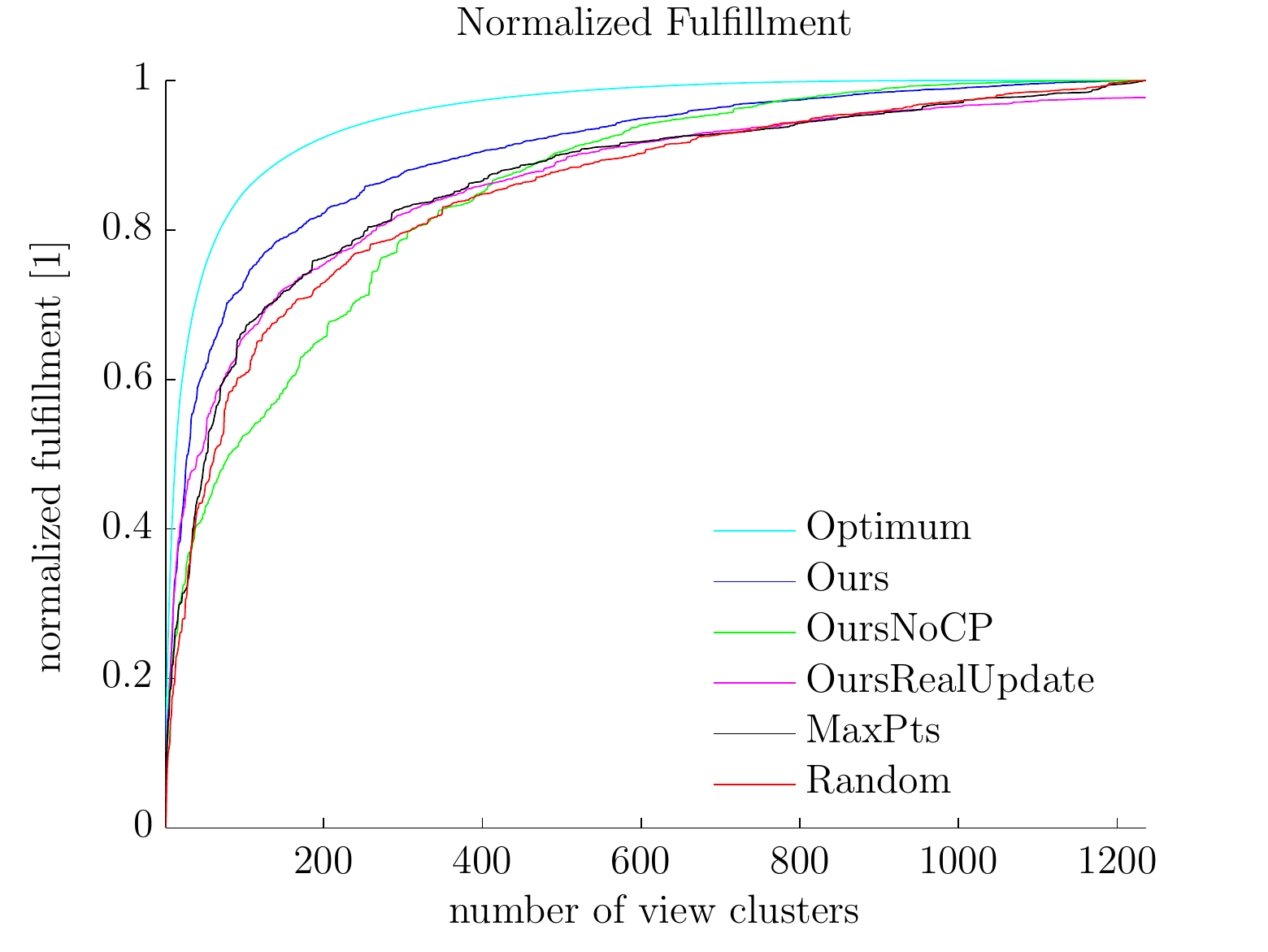} 
}
\quad
\hspace{-40pt}
 \subfigure[House]
    {
                \includegraphics[width=0.5\columnwidth]{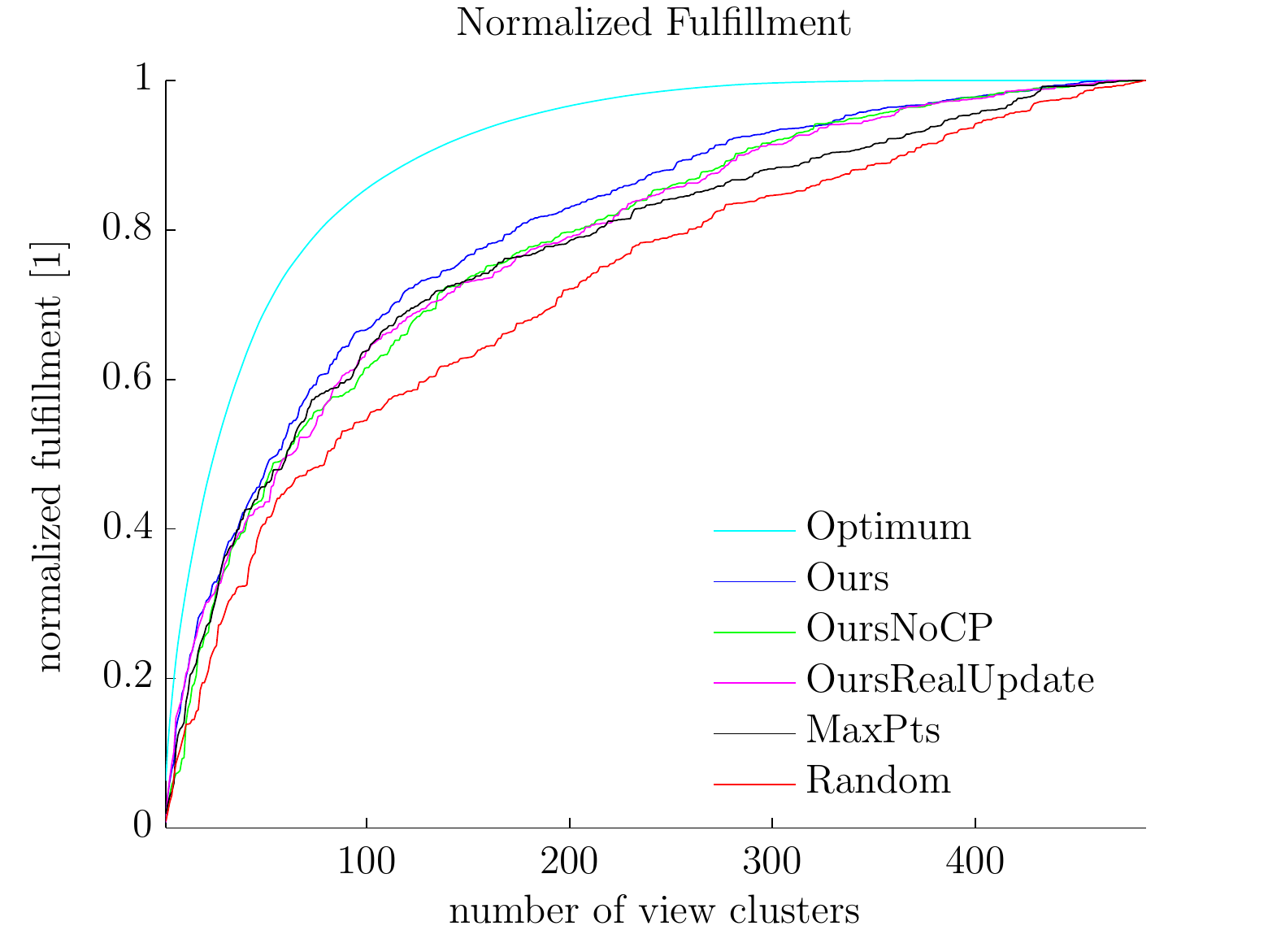} 
}
    \caption{Ranking Performance on the {\bf Valley}  and {\bf House}  Datasets. We show the normalized fulfillment
    over the number of used view clusters. 
    }
  \label{fig:ranking_valley}
\end{figure}

\begin{table}
\centering
 \begin{tabular}[b]{|c||c||c|c|c|c|c|}
  \hline
   Fulfillment  &  Opt &  Ours  &   OursNoCP & OursRU &  MaxPts  &  Random \\\hline 
   Threshold  &  \multicolumn{6}{c|}{Required View Clusters [\%]}  \\\hline\hline 
    10\% &  0.08 & \bf 0.24 & \bf 0.24 & \bf 0.24 & \bf 0.24& 0.32\\\hline  
    20\% &  0.24 &  0.57 & \bf 0.49 & 0.57 & 0.73 & 1.05\\\hline     
    30\% &  0.49 &  0.97 & 1.46 & \bf 0.89 & 1.62 & 2.18\\\hline       
    40\% &  0.73 & \bf 1.46 & 3.07 &  1.62 & 2.83 & 2.99\\\hline        
    50\% &  1.13 & \bf 2.43 & 6.63 & 3.72 & 4.21 & 5.02\\\hline  
    60\% &  1.78 & \bf 4.37 & 12.78 & 5.99 & 6.07 & 7.36\\\hline 
    70\% &3.16 & \bf 7.12 & 19.17 & 10.92 & 10.60 & 13.03\\\hline 
    80\% &5.66 & \bf 14.00 & 25.16 & 21.60 & 20.71 & 25.16\\\hline 
    90\% &12.70 & \bf 30.91 & 39.56 & 42.07 & 39.97 & 47.98\\\hline 
\end{tabular}
\caption{View Cluster Ranking on the Valley Dataset. We show the percentage of view clusters that are required for reaching a certain percentage
of the maximum achievable fulfillment. Note that our approach only require two times the optimal number
of view clusters, whereas the second best approach requires between 30-90\% more than ours.
 }
\label{tab:ranking}
\end{table}

In this experiment, we evaluate our approach with prediction (Ours) against two variants of our approach.
The first variant (OursNoCP) does not use the confidence prediction (i.e. $f_{conf}$ is fixed to 1).
The second variant (OursRealUpdate) uses the confidence prediction, but instead of also predicting the 
update of the objective function, it uses the real fulfillment for the update.
This variant can be seen as our approach with the MVS approach in the loop.

Additionally to random ranking (Random), we also tried to come up with a good baseline algorithm that only operates on the sparse point cloud.
We denote this algorithm as "MaxPts".
This algorithm always greedily selects the view cluster of the camera with the highest number of connected 3D points.
After selecting the camera with the maximum number of points, all these points are removed from the sparse reconstruction.
Through this removal operation, the algorithm naturally tries to explore the reconstruction.
 Finally, we also compare our approach to the optimal greedy solution (Optimum).
 This algorithm requires all reconstruction to be available in every iteration and
 uses the real objective function for all its decisions.

\paragraph{Results}
In Figure~\ref{fig:ranking_valley}, we show the normalized fulfillment for all approaches, such that the highest fulfillment value is set to one.
In Table~\ref{tab:ranking}, we also show the percentage of view clusters, which are needed to reach a certain fulfillment
level in steps of 10\%.

If we analyze the results of the Valley Dataset, we can see the following.
First of all, our full approach performs very well. 
Between 30\% and 70\% (which is one of the most interesting regions for our task),
the second best approach requires approximately 70\% more view clusters than our approach.
What seems really fascinating at first glance is that in this region our approach with prediction (Ours)
actual performs better than with the MVS algorithm in the loop (OursRealUpdate).
The reason for this astonishing result is that, with the real update, the algorithm does not have any notion 
of what it has tried in the past.
This means, if for some reason the prediction says that there is a chance to reconstruct this object from this view point
and the reconstruction actually fails, the algorithm will try the same thing with the view directly next to the last one.
However, with the prediction such a behavior is naturally avoided.
If we completely remove the prediction from our approach (OursNoCP), the performance degrades very soon below the baseline method.
If we take a closer look at Table~\ref{tab:ranking}, we can see that the first 60\% of fulfillment are quite cheap with our approach.
For 60\% we only require 4.4\% of all poses (i.e. 45 view clusters). However, each further added 10\% of fulfillment
roughly doubles the required number of view clusters.

If we analyze the results of the House Dataset (Figure~\ref{fig:ranking_valley}),
we can see that our approach still performs better than the baseline approaches from 40\% upwards.

\subsection{Full View Cluster Ranking}
\label{ss:full}
In the previous two experiments, we analyzed each of our two steps separately.
In this experiment, we evaluate our full approach against the best combination
of the baseline methods.
We denote this approach as "Max",
as it consists of the matching partner selection based on the maximum connectivity (MaxCon)
and the ranking procedure based on the maximum number of visible 3D points (MaxPts).
For both approaches, we evaluate the performance for different numbers of matching partners,
i.e. $k = {2,3,5,11}$.

\begin{figure}[tp]
  \centering
 \subfigure[Valley]
    {
                \includegraphics[width=0.5\columnwidth]{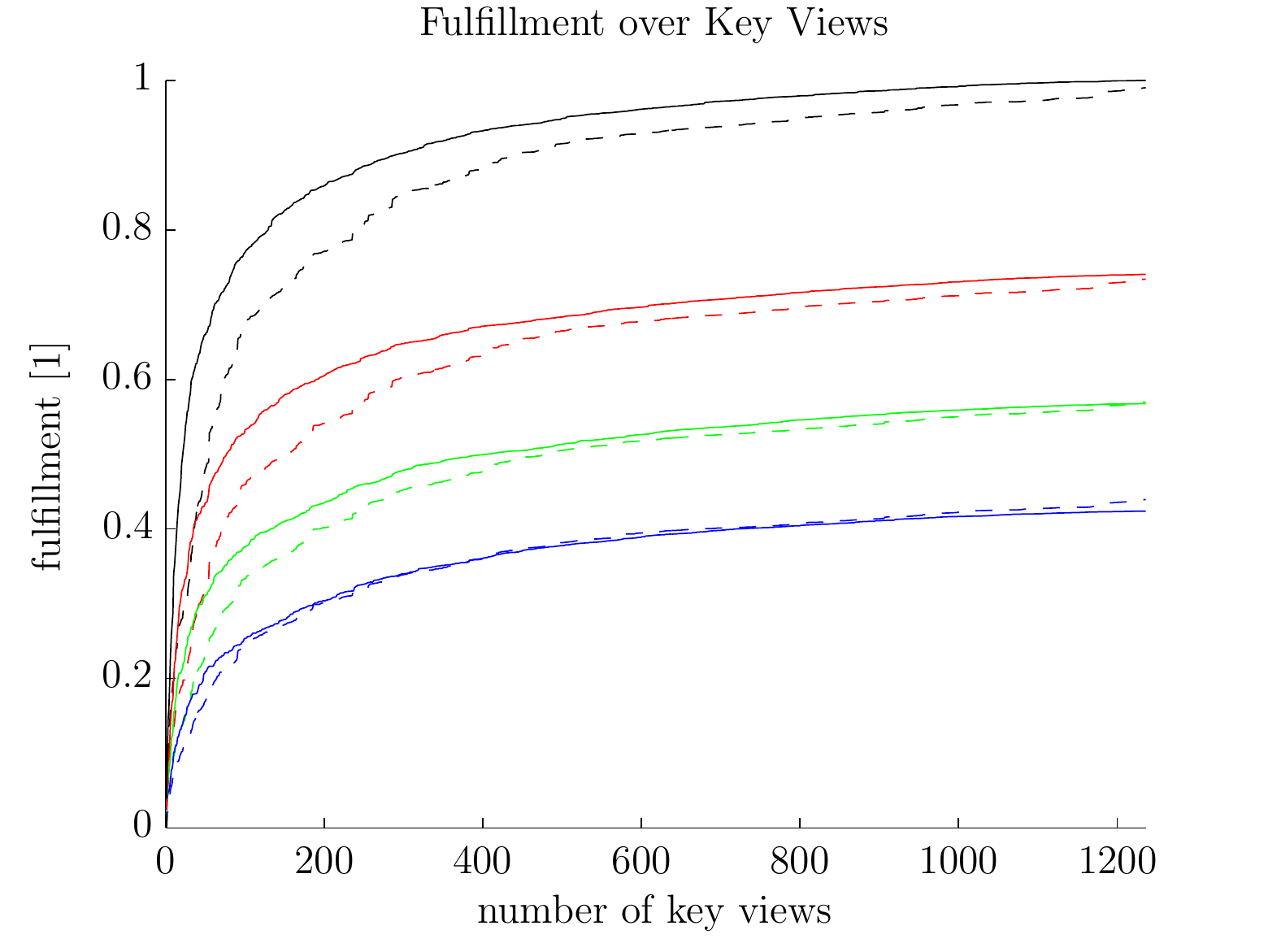} 
}
\quad
\hspace{-40pt}
 \subfigure[Valley]
    {
                \includegraphics[width=0.5\columnwidth]{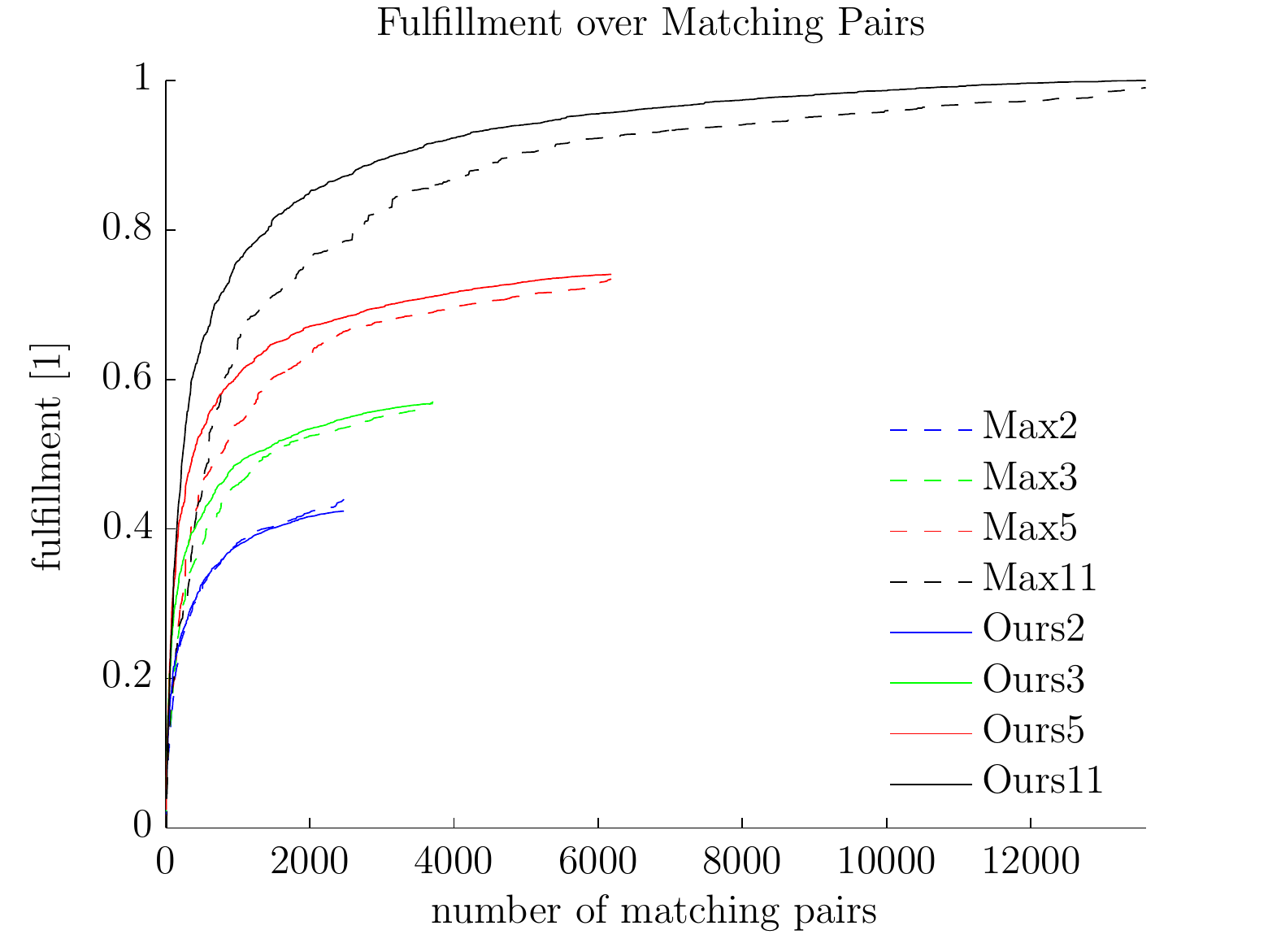} 
} \quad
 \subfigure[House]
    {
                \includegraphics[width=0.5\columnwidth]{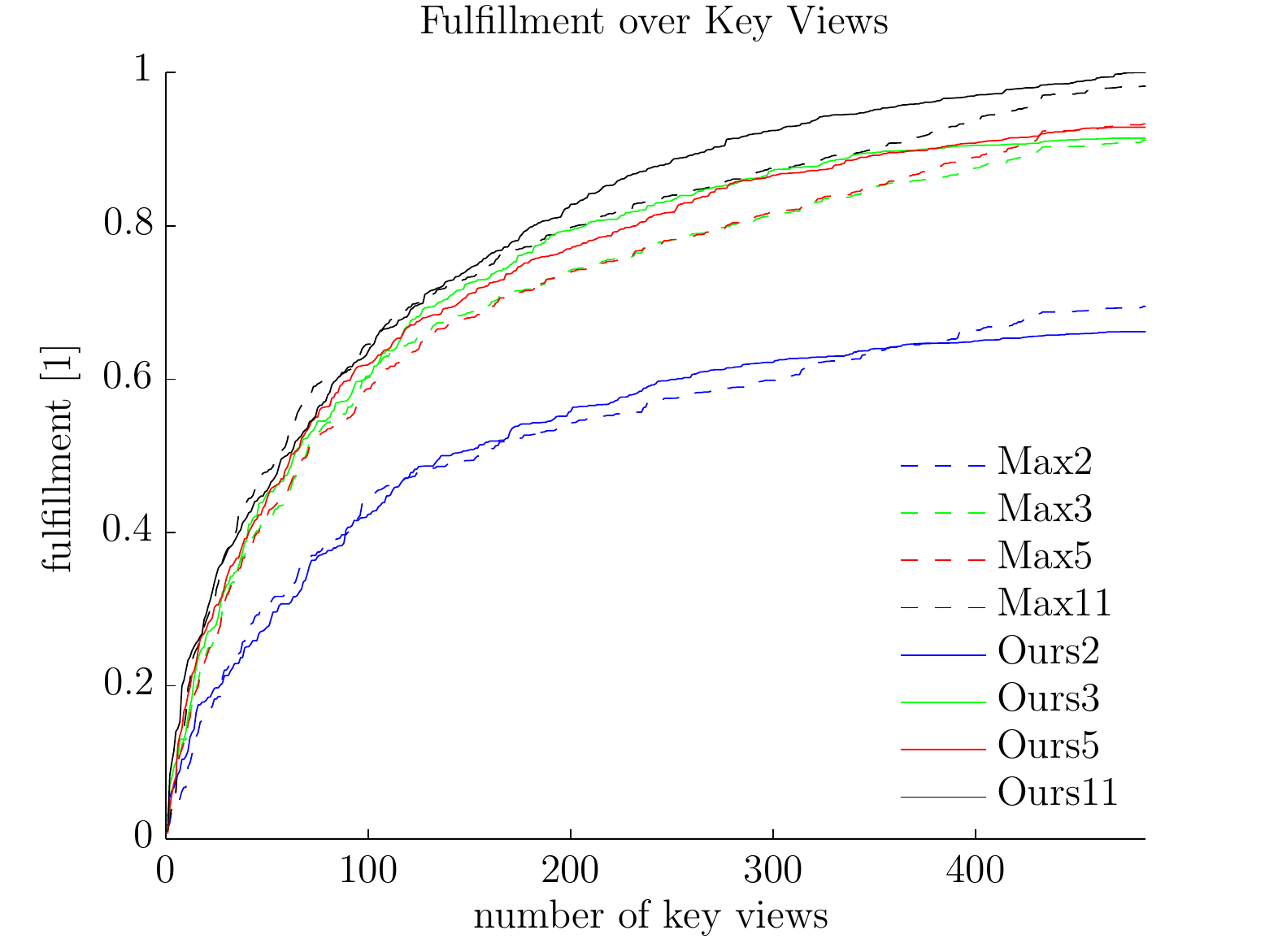} 
}
\quad
\hspace{-40pt}
 \subfigure[House]
    {
                \includegraphics[width=0.5\columnwidth]{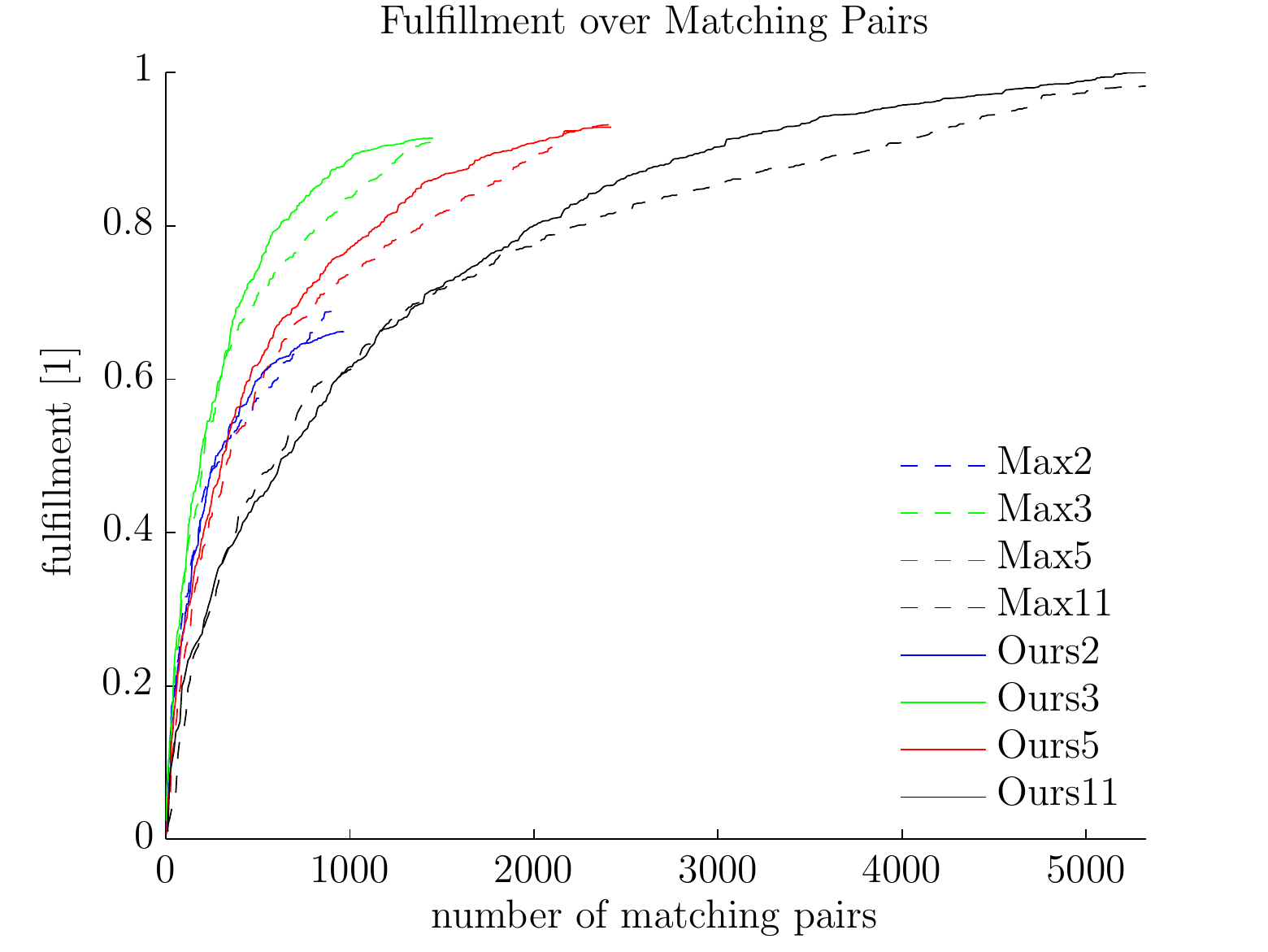} 
}
    \caption{Fulfillment development of our full approach on the  {\bf Valley Dataset} (top) and {\bf House Dataset} (bottom). We show our full approach (Ours) and the baseline method (Max)
    for four different numbers of matching partners (i.e. $k = {2,3,5,11}$). The left side shows the fulfillment over the number of key views,
    whereas the right side shows the fulfillment over the number of matching pairs. 
    }
  \label{fig:full_valley}
\end{figure}

\begin{figure}[tp]
  \centering
 \subfigure[Valley]
    {
                \includegraphics[width=0.5\columnwidth]{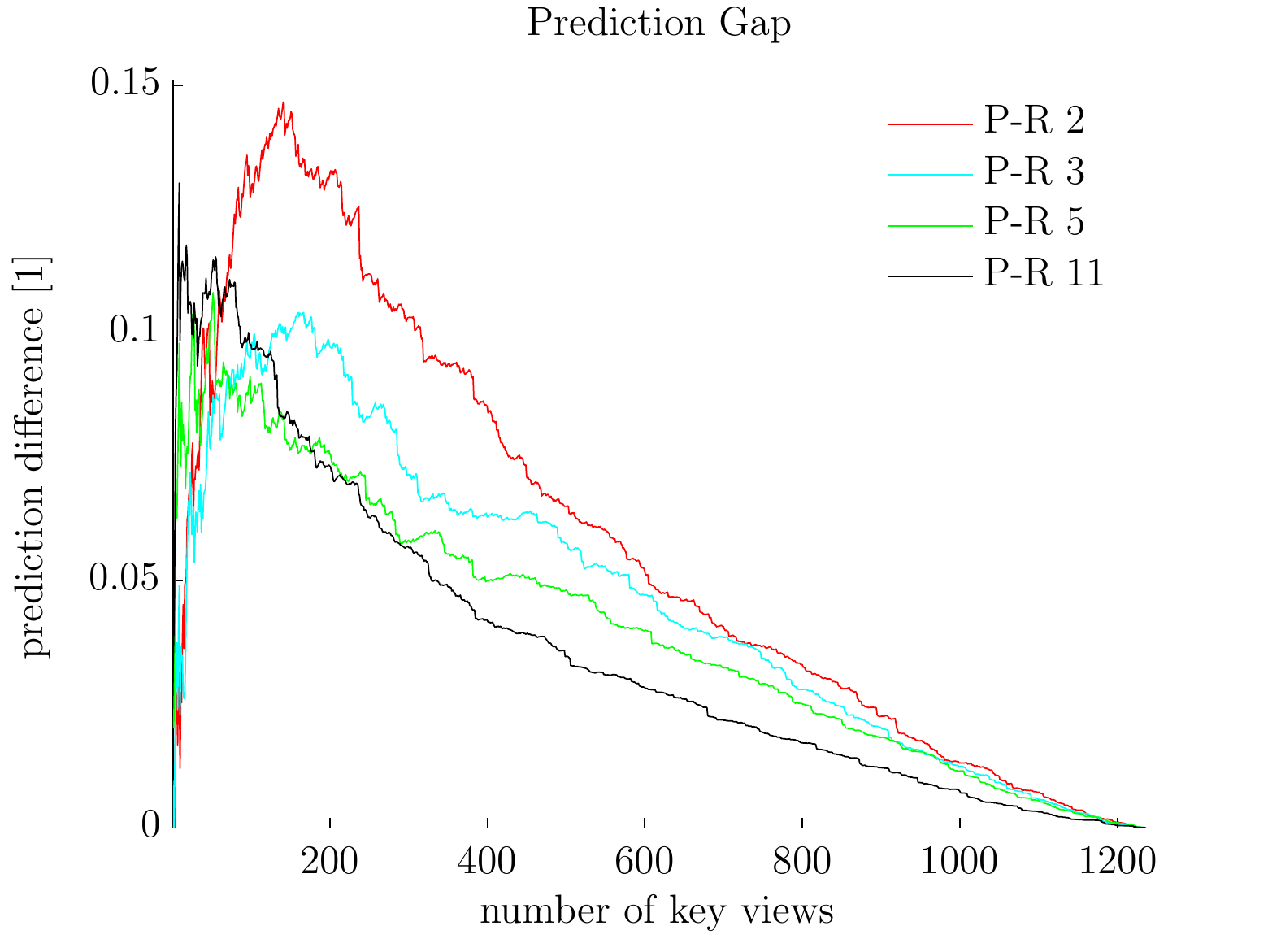} 
}\quad
\hspace{-40pt}
 \subfigure[House]
    {
                \includegraphics[width=0.5\columnwidth]{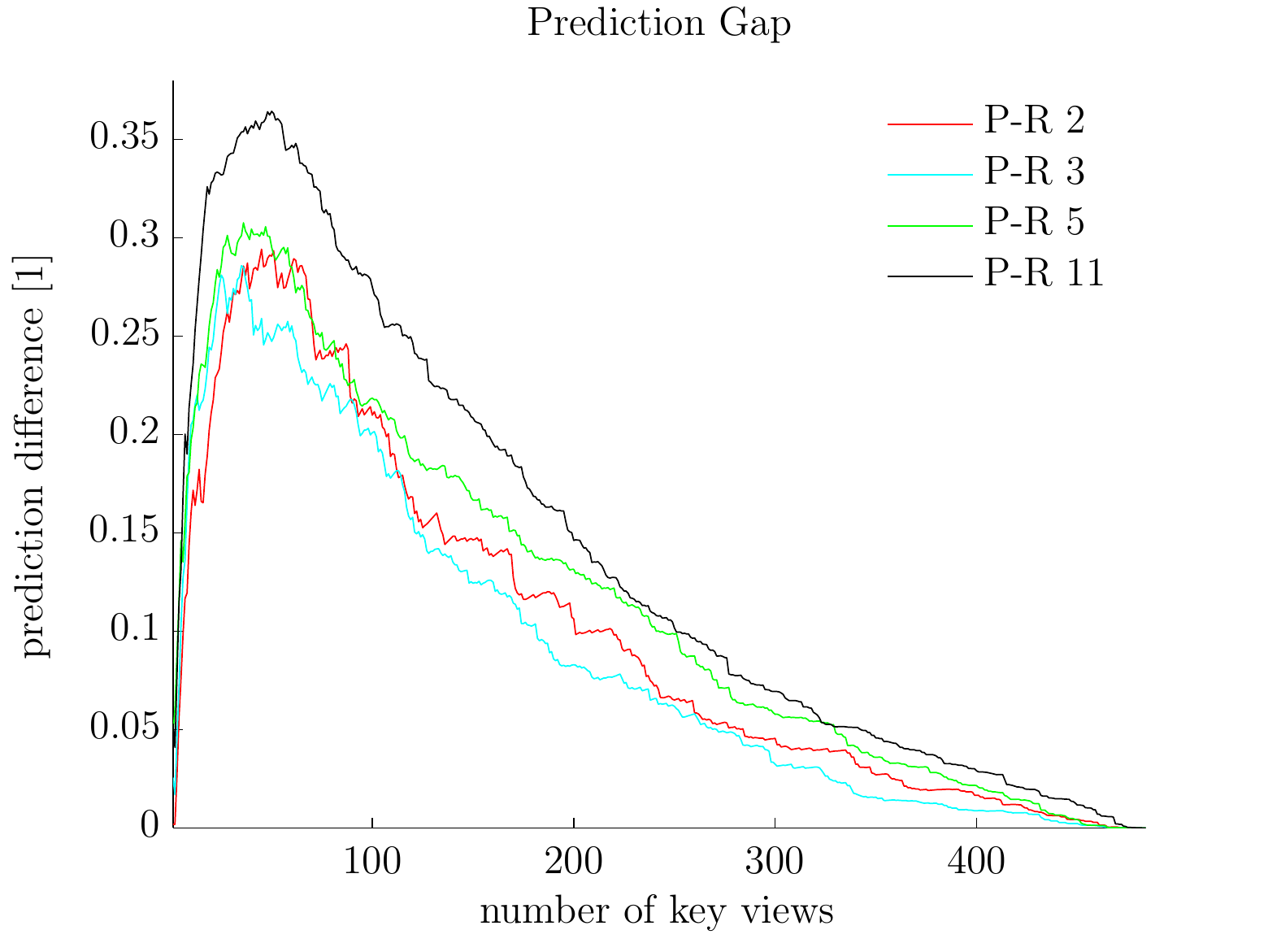} 
}
    \caption{
    We show the prediction gap as the difference of the normalized fulfillment (i.e. Predicted minus Real) for the {\bf Valley} and {\bf House} Dataset. 
    Note that the prediction gap on the Valley dataset is not very wide (especially for 5 and 11 matching partners).
    }
  \label{fig:real_vs_pre}
\end{figure}

\paragraph{Results}
In Figure~\ref{fig:full_valley}, we show the fulfillment curves for both datasets.
Note that the curves are normalized such that maximally achieved fulfillment over all approaches (i.e. Ours11 with all key views)
is set to one. 
On a first glance, we can see that there is a significant difference in the datasets.
For the Valley Dataset, there is a significant difference in the maximally reached fulfillment
for different numbers of matching partners.
E.g. our approach with 3 matching partners is only able to reach half of the fulfillment of the 
approach with the maximum number of matching partners.
For the House Dataset this gap is significantly smaller and our approach with 3 matching partners is able to reach
90\% of the fulfillment.
We think that the main reason for this discrepancy is the difficulty of the dataset.
While the Valley Dataset is strongly dominated by trees which are exceedingly hard to reconstruct,
the House Dataset contains mostly flat structures such as roads, short grass or roofs.

For the Valley Dataset, our approach significantly outperforms right from the start.
In Table~\ref{tab:full_valley}, we see that 
between 30\% and 70\% fulfillment the baseline approach with 11 matching partners requires 
2 to 3 times as many matching pairs than our approach for obtaining the same fulfillment level.

If we take a look at the House Dataset,
we can also see a clear gap between our approach and the baseline.
However, this clear gap only starts to form between 60\% and 70\% fulfillment.
What seems interesting is that in terms of computational efficiency, 
three matching partners seem to be the best choice for the House Dataset.

If we take a look at the prediction performance (Figure~\ref{fig:real_vs_pre}),
we see a significant difference between the training environment (Valley) and the unseen test environment (House).
While there is a significant gap between predicted fulfillment and real fulfillment on the House Dataset,
the gap is a lot smaller on the Valley Dataset.
In fact, the gap closes below 10\% after only 55 key views for 5 matching partners (87 for 11).
This means in a known environment, the predicted fulfillment can indeed be used for estimating the actual fulfillment
before executing the actual MVS algorithm.

\begin{table}
\centering
 \begin{tabular}[b]{|c||c|c|c|c|c|c|c|c|}
  
  \cline{2-9}
   \multicolumn{1}{c|}{}&  \multicolumn{4}{c|}{Max} & \multicolumn{4}{c|}{Ours} \\\cline{2-9}
    \multicolumn{1}{c|}{} &  2 & 3  & 5& 11 & 2 & 3  & 5& 11   \\\hline 
   Fulfillment  &  \multicolumn{8}{c|}{Required Matching Pairs [\%]}  \\\hline\hline 
    10\% &0.29 & 0.26 & 0.22 & 0.24 & \bf 0.15 &\bf  0.15 & 0.18 & 0.16\\\hline  
    20\% &0.94 & 0.82 & 0.92 & 0.89 & 0.71 & \bf 0.35 & 0.40 & 0.49 \\\hline     
    30\% &2.88 & 1.83 & 1.54 & 2.18 & 2.77 & 1.04 & \bf 0.70 & 0.81 \\\hline       
    40\% &9.91 & 4.30 & 2.57 & 2.99 & 10.61 & 2.91 & 1.29 & \bf 1.13 \\\hline        
    50\%  &- & 10.55 & 5.63 & 4.45 & - & 8.94 & 2.83 & \bf 1.78 \\\hline  
    60\%  &- & - & 10.70 & 5.99 & - & - & 7.02 &\bf  2.67 \\\hline 
    70\%  &- & - & 30.78 & 10.03 & - & - & 22.76 &\bf  5.02 \\\hline 
    80\%&- & - & - & 19.26 & - & - & - & \bf 10.52\\\hline 
    90\% &- & - & - & 36.25 & - & - & - &\bf  23.46 \\\hline 
\end{tabular}
\caption{Required number of matching pairs to reach a certain level of fulfillment on the {\bf Valley Dataset }.
We compare our full approach (Ours) to the best baseline combination (Max) for a varying number of matching partners.
The best values of each row are marked in bold font.
 }
\label{tab:full_valley}
\end{table}

\subsection{Discussion}
\label{ss:discussion}
In all our experiments, we have shown that the MVS confidence prediction allows us to consistently outperform the exact same approach without prediction.
We see this as proof that our prioritization approach was able to use the accumulated knowledge of the confidence predictor to significantly improve the performance.
In this sense, we think that our approach could be a great benefit to reoccurring photogrammetric tasks 
in difficult environments.

On the Valley Dataset (i.e. the dataset which was captured in the same domain as the training data),
we were able to achieve significantly better results for the view cluster prioritization than any other baseline.
For the same level of fulfillment our approach requires 2 to 3 times less key views than the best baseline.
Further, we demonstrated that the real fulfillment only lacks approximately 10\% behind the predicted fulfillment.
This means that the fulfillment prediction is a reasonable approach for approximating the necessary number 
of key views to get a good coverage of the scene.
E.g. if we aim for 70\% fulfillment with 11 matching partners, we can use our approach to predict a fulfillment of 80\%.
In this experiment, this method would lead to 69.3\% fulfillment with only 60 key views (which is already highly complete -- see video).
If we analyze the computation time for this example in relation to using each view as key view (as done by conventional MVS),
we end up with a speed up factor of 9.2
(from $n / (s+n \cdot r)$ with number selected of key views $n$, number selected of key views $s$ and the ratio $r$ of prioritization time over MVS time).
If we are only interested in the saved memory, we even achieve a memory reduction factor of 20.1 (430M points vs. 8650M points).
This means that our approach has a large potential to save computational time and memory with a very small overhead.

Our experiments on the House dataset
 demonstrated that our approach still performs reasonable even if confronted with a scene that 
contains mostly objects that were never seen in training.
Our approach still has some benefit over the baseline methods for a high degree of fulfillment,
however, the margin is significantly less than for the Valley dataset.
On the House dataset, we reach 70\% of the achievable fulfillment with
7.6\% of the available matching pairs, while the 
corresponding baseline requires 9.1\%.
However, if we compare our the actual run-time to the best baseline,
we see that our method does not lead to a run-time improvement for the same fulfillment level
due to the higher computational overhead.
In this sense, we can conclude that our method will not break down in an unknown environment,
but the main benefit over non-learning approaches will be significantly diminished.

\section{Conclusion}
In this work, we presented a new approach to prioritize view clusters (key views with matching partners) for obtaining
a complete and accurate 3D reconstruction at a low computational time and memory consumption.
The proposed approach uses unsupervised machine learning for accumulating empirical evidence under what circumstances 
an MVS algorithm does not work as intended.
In the prioritization, we use this accumulated knowledge to prefer constellations with a high probability of success and 
thus maximize the reconstruction completeness.
In our experiments, we demonstrate that our approach is thus able to reach a certain level of quality fulfillment
(in terms of completeness with respect to a desired accuracy and ground resolution) with up to three 
times less key views than the best baseline approach.
In contrast to other approaches, 
our approach has the distinct advantage that it provides the opportunity to predict the 
relative fulfillment level in complex scenes.
This means that the user can decide whether doubling the amount of computation time and memory consumption
for reaching 80\% fulfillment instead of 70\% (which is hardly visible -- see video) is necessary or not.
On the topic of domain generalization, we have shown that our approach still performs reasonably well
in an environment that has never been seen in training,
however, the advantage over non-learning based approaches is significantly diminished.
In this sense, we see the main application area of our approach in reoccurring photogrammetric tasks
in challenging environments.

\section*{Acknowledgements}\label{acknowledgements}
Results incorporated in this paper received funding from the European Unions Horizon 2020 research and innovation programme under
grant agreement No 730294 and the EC FP7 project 3D-PITOTI (ICT-2011-600545).

\section*{References}

\bibliography{\BibPath/icg_abbrevs,\BibPath/user-bibtex}

\appendix
\section{Probabilistic Confidence Extension}
\label{conf_extension_appendix}
\begin{figure}[t]
  \centering
\includegraphics[width=0.6\columnwidth]{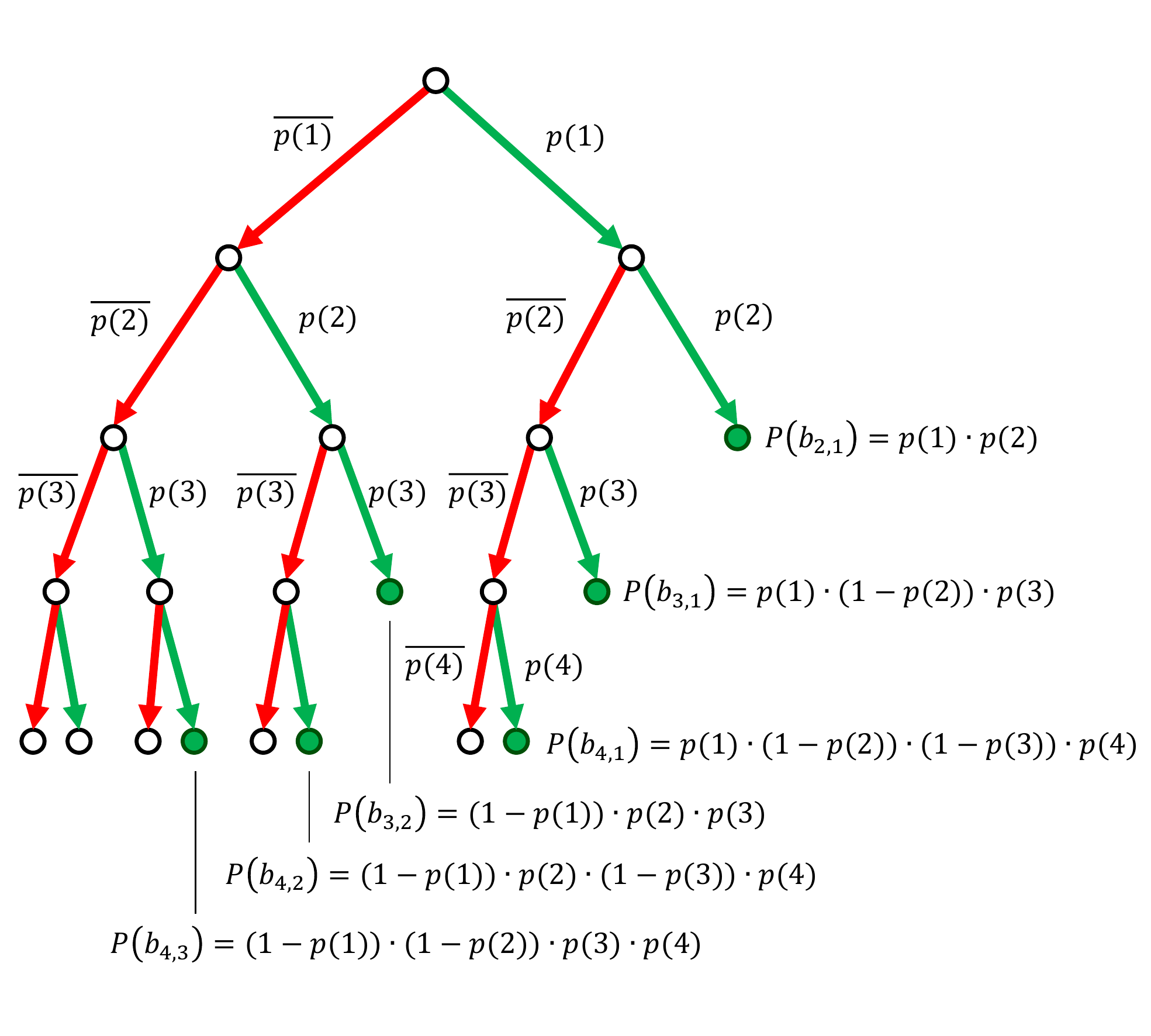} 
    
  \caption{Probability tree. We formulate the MVS confidence for a key view with $k$ cameras as matching partners as a stochastic process (here $k=4$). 
  Each edge in the tree relates to an event. For a matching partner $d$, the event can either be positive (green color) with a probability $p(d)$ or
  negative (red color) with a probability $\overline{p(d)} = 1 - p(d)$. A positive event corresponds to a successful match between the matching partner $d$ and the key view at a specific pixel location,
  which results in a 3D measurement that conforms with our uncertainty model. For obtaining a reliable 3D reconstruction, we require at least two matching partners with a successful match.
  Thus, each path in the tree with at least two positive events can be seen a successful branch. Each successful branch $b_{y,x}$ is defined by the first 
  two positive events at level $x$ and $y$ respectively.
  The overall probability of a successful measurement can be obtained by summing all successful branches.
  Note that the order events in the tree is irrelevant for the overall probability, which can be better seen in Equation~\ref{eq:pmin2}.
  }
  \label{fig:prob_tree}
\end{figure}

 For extending the confidence prediction framework to an arbitrary number of matching partners,
 we use a traditional probability tree. 
 In Figure~\ref{fig:prob_tree} we show such a probability tree for 4 matching partners.
 In the following, we use such a general probability tree to derive a probabilistic formulation for obtaining a successful 3D measurement 
from $k$ cameras.
 
Let us first formalize the probability tree for $k$ cameras as a binary tree with a depth $k$.
Every depth level $d$ represent more or less what happens if we add a $d^{th}$ camera to the previous set of $d-1$ cameras.
Every vertex in the tree represents an event, which is successful with a probability $p(d)$,
where $d$ is the depth of the tree (also corresponding to the $d^{th}$ matching partner).
Let us call a path from the root vertex to a leaf vertex (i.e. a vertex without child vertices) simply "branch".

Now let us grow the binary tree.
As we are only interested in having at least 2 successful matches, we can stop growing a branch when it has reached
two positive events (we will further call this kind of branch as "successful branch").
As such a successful branch has always exactly two successful events, we will further denote a successful branch as $b_{y,x}$,
where $y$ is the depth/id of the last successful event and $x$ is the depth/id of the preceding successful event.
This means that only branches with less than 2 successful events are grown in the next depth level.
It follows that in the process of growing from level $d-1$ to $d$ (i.e. adding a $d^{th}$ camera),
the tree gains exactly $d-1$ new successful branches (one for each camera already in the tree),
while all old successful branches remain unchanged.
It also follows that
each new successful branch ($b_{d,i}$) has exactly one previous camera $i$ (with $1 \geq i < d$) with a successful event associated.
Thus the probability along such a new successful branch $b_{d,i}$ is given by:
\begin{equation}
 P(b_{d,i}) = p(d) \cdot p(i) \prod_{j \in \{1,2,...,d-1\} \backslash i } (1 - p(j)),
\end{equation}
where $p(x)$ is the probability of a successful match with camera $x$.
If we now add up all branches with 2 successful events at level $d$, we end up with the following probability
for obtaining at least 2 successful events:
\begin{equation}
 P_{min 2} =  \sum_{i=2}^d \sum_{a=1}^{i-1} P(b_{i,a}) = \sum_{i=2}^d \sum_{a=1}^{i-1} p(i) \cdot p(a) \prod_{j \in \{1,2,...,i-1\} \backslash a } (1 - p(j))
\end{equation}
Now we can expand the equation to
\begin{equation}
\begin{split}
 P_{min 2}  =  \sum_{i=2}^d  p(i) \cdot \left(  \left(\sum_{c=1}^{i-1} p(c) \right) -  2 \left( \sum_{c^{(2)} \in C^{(2)}_{i-1}} p\left( c^{(2)}_1\right) \cdot p\left(c^{(2)}_2 \right)  \right) \right. \\
 \left. + 3 \left( \sum_{c^{(3)} \in C^{(3)}_{i-1}} \prod_{a=1}^3 p\left(c^{(3)}_a \right)  \right) - ... + (-1)^{i} (i-1) \left( \sum_{c^{(i-1)} \in C^{(i-1)}_{i-1}} \prod_{a=1}^{i-1} p\left(c^{(i-1)}_a \right)  \right)  \right) ,
\end{split}
 \end{equation}
 where $C^{(x)}_{i-1}$ is the solution space for drawing subsets of $x$ cameras
from the available set of ${i-1}$ cameras, $c^{(x)}$ is one of these subsets and 
$c^{(x)}_a$ is one camera of this subset.
Now we can contract the equation again to
 \begin{equation}
 P_{min 2} =  \sum_{i=2}^d \left( (-1)^i \cdot (i-1) \cdot \sum_{c^{(i)} \in C^{(i)}_d }  \left( \prod_{a=1}^i p\left(c^{(i)}_a \right)  \right) \right) 
 \label{eq:pmin2}
\end{equation}
Note that this equation and Equation~\ref{eq:prob} are equivalent and that only the parameterization was changed;
i.e.  $d \mapsto  k$,
 $P_{min 2} \mapsto f_{conf}(t,c_{key}, c^k)$ and  $p\left(c^{(i)}_a \right) \mapsto  f_{conf}\left(c_{key},c^{(i)}_a\right) $.

\end{document}